\documentclass[10pt,twocolumn,letterpaper]{article}

\usepackage{iccv}              

\usepackage{graphicx}
\usepackage{amsmath}
\usepackage{amssymb}
\usepackage{booktabs}
\usepackage{pifont}
\usepackage{threeparttable} 
\usepackage[utf8]{inputenc}
\usepackage{float}
\usepackage{pifont}
\usepackage{footnote}
\usepackage{enumitem}
\usepackage{bm}
\usepackage{arydshln}
\usepackage{booktabs}
\usepackage{multicol}
\usepackage{multirow}
\usepackage{color}
\usepackage{xcolor}     
\usepackage{colortbl}
\usepackage{soul}
\usepackage{bbding}
\usepackage{makecell}
\usepackage{mathtools}
\usepackage{imakeidx}
\makeindex
\usepackage{arydshln}
\usepackage{lipsum}
\usepackage{natbib}
\usepackage[toc]{multitoc}
\usepackage[edges]{forest}
\usepackage[normalem]{ulem}

\usepackage{bbding}
\usepackage[most]{tcolorbox}

\usepackage{algorithm}
\usepackage{algorithmic}

\definecolor{orchid}{rgb}{0.85, 0.44, 0.84}
\definecolor{iblue}{rgb}{0.06, 0.75, 1.0}
\definecolor{igray}{rgb}{0.00, 0.00, 0.00}
\definecolor{ired}{rgb}{0.8588, 0.2666, 0.2156}

\newcommand{\ourdataset}{{\fontfamily{ppl}\selectfont AbdomenAtlas 2.0}}
\newcommand{\jhhdataset}{proprietary}
\newcommand{\ourmodel}{{\fontfamily{ppl}\selectfont Tumor Genesis}}
\newcommand{\ourtool}{SMART-Annotator}

\newcommand{\lowerbound}{500}
\newcommand{\upperbound}{1,500}
\newcommand{\fullnumber}{3,000}
\newcommand{\numofscans}{10,135}
\newcommand{\numoftumorinst}{15,130}
\newcommand{\numofcontrolscans}{5,893}

\newcommand{\numofradiologists}{23}

\newcolumntype{P}[1]{>{\centering\arraybackslash}p{#1}}
\newlength\savewidth\newcommand\shline{\noalign{\global\savewidth\arrayrulewidth
  \global\arrayrulewidth 0.8pt}\hline\noalign{\global\arrayrulewidth\savewidth}}

\definecolor{iccvblue}{rgb}{0.21,0.49,0.74}
\usepackage[pagebackref,breaklinks,colorlinks,allcolors=iccvblue]{hyperref}

\def\paperID{4977} 
\def\confName{ICCV}
\def\confYear{2025}

\title{Scaling Tumor Segmentation:\\Best Lessons from Real and Synthetic Data}

\author{
Qi Chen\textsuperscript{1,2} \quad 
Xinze Zhou\textsuperscript{1} \quad
Chen Liu\textsuperscript{1,3} \quad
Hao Chen\textsuperscript{4} \quad 
Wenxuan Li\textsuperscript{1} \quad 
Zekun Jiang\textsuperscript{5} \\
Ziyan Huang\textsuperscript{6,7} \quad
Yuxuan Zhao\textsuperscript{8} \quad
Dexin Yu\textsuperscript{8} \quad
Junjun He\textsuperscript{7} \quad 
Yefeng Zheng\textsuperscript{9} \quad 
Ling Shao\textsuperscript{2} \\
Alan Yuille\textsuperscript{1} \quad 
Zongwei Zhou\textsuperscript{1,}\thanks{Correspondence to Zongwei Zhou (\href{mailto:zzhou82@jh.edu}{\textsc{zzhou82@jh.edu}})} \\ [2.5mm]
\textsuperscript{1}Johns Hopkins University \quad \\
\textsuperscript{2}UCAS-Terminus AI Lab, University of Chinese Academy of Sciences \quad \\
\textsuperscript{3}Hong Kong Polytechnic University \quad 
\textsuperscript{4}University of Cambridge \quad \\
\textsuperscript{5}Sichuan University \quad 
\textsuperscript{6}Shanghai Jiao Tong University \quad  \\
\textsuperscript{7}Shanghai AI Laboratory \quad 
\textsuperscript{8}Qilu Hospital of Shandong University \quad
\textsuperscript{9}Westlake University 
\\ [1.5mm]
{\small Code, Model \& Data: ~\href{https://github.com/BodyMaps/AbdomenAtlas2.0}{https://github.com/BodyMaps/AbdomenAtlas2.0}}
}

\begin{document}
\maketitle

\begin{abstract}

AI for tumor segmentation is limited by the lack of large, voxel-wise annotated datasets, which are hard to create and require medical experts. In our proprietary JHH dataset of \textbf{\fullnumber} annotated pancreatic tumor scans, we found that AI performance stopped improving after \textbf{\upperbound} scans. With synthetic data, we reached the same performance using only \textbf{\lowerbound} real scans. This finding suggests that synthetic data can steepen data scaling laws, enabling more efficient model training than real data alone.
Motivated by these lessons, we created \textbf{\ourdataset}---a dataset of \numofscans\ CT scans with a total of \numoftumorinst\ tumor instances per-voxel manually annotated in six organs (pancreas, liver, kidney, colon, esophagus, and uterus) and \numofcontrolscans\ control scans. Annotated by \numofradiologists\ expert radiologists, it is several orders of magnitude larger than existing public tumor datasets. While we continue expanding the dataset, the current version of \ourdataset\ already provides a strong foundation---based on lessons from the JHH dataset---for training AI to segment tumors in six organs. It achieves notable improvements over public datasets, with a \textbf{+7\%} DSC gain on in-distribution tests and \textbf{+16\%} on out-of-distribution tests.

\end{abstract}

\section{Introduction}\label{sec:intro}

\begin{figure}[t]
	\centering
	\includegraphics[width=\linewidth]{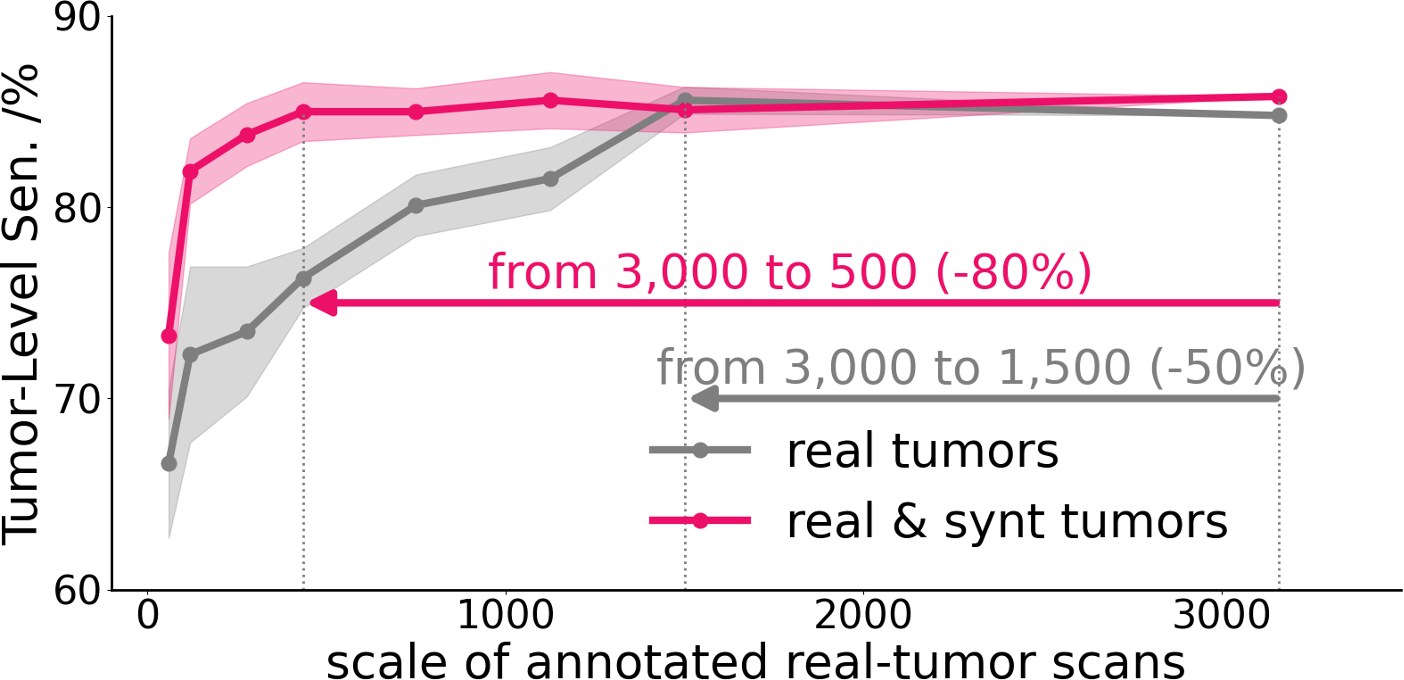}
    \caption{\textbf{Data scaling laws study.} Experimental results on the  \jhhdataset~dataset demonstrate that increasing the scale of real data  improve the segmentation  (gray curve). Notably, supplementing the dataset with an additional $3\times$ synthetic data (red curve) can further enhance the results, revealing  the potential of a larger public dataset to advance tumor research.
     }
\label{fig:teaser}
\end{figure}

\begin{figure*}[t]
	\centering
	\includegraphics[width=\linewidth]{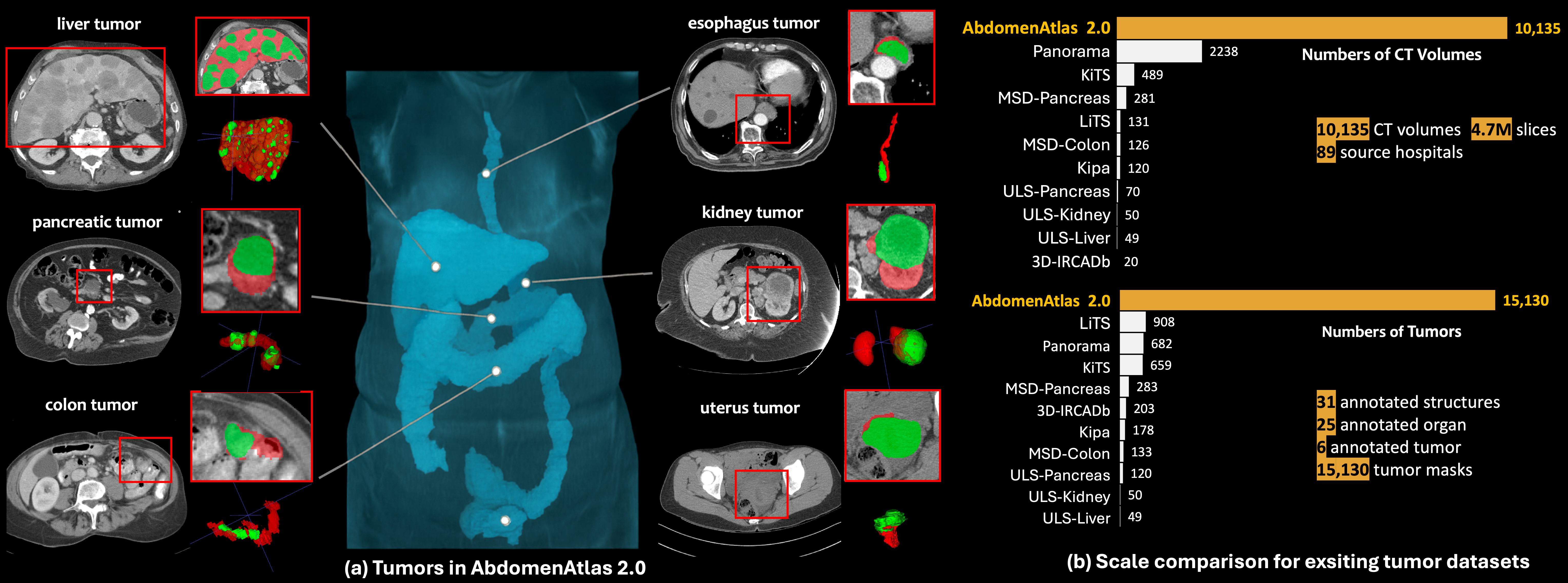}
	\caption{\textbf{Overview of the \ourdataset\ dataset.} For each CT scan,  \ourdataset\ provides precise and high-quality annotations following a well-designed AI-driven annotation pipeline.  Compared to existing datasets, \ourdataset\ collects large-scale CT scans from diverse clinical sources, encompassing a wide range of tumor types (\ie, liver, pancreas, kidney, colon, esophageal, and uterine tumors) and comprehensive tumor sizes.  This extensive scale makes it the largest human-annotated tumor mask dataset. 
    }
	\label{fig:ourdataset}
\end{figure*}

Developing AI models for tumor segmentation is fundamentally challenged by the scarcity of large, annotated datasets---owing to the immense time and expertise required for per-voxel annotation \cite{ma2024fast,zhou2021towards,zhou2022interpreting}. Inspired by scaling laws \cite{kaplan2020scaling,rombach2022high,ouyang2022training,fan2024scaling}, to estimate the impact of data scale on tumor segmentation performance, we first leveraged a proprietary dataset of \fullnumber\ pancreatic tumor scans, per-voxel annotated over five years by expert radiologists and verified by pathology reports. Our previous work~\cite{xia2022felix,li2025scalemai} showed that this dataset enabled AI to reach radiologist-level detection accuracy. However, as shown in~\figureautorefname~\ref{fig:teaser}, performance gains plateaued after \upperbound\ scans, suggesting diminishing returns from adding more real data. Recognizing that annotating \upperbound\ scans is still a considerable undertaking for a single tumor type, we explored the potential of synthetic data \cite{hu2022synthetic,li2023early,chen2024analyzing,li2024text,mao2025medsegfactory,guo2025text2ct} to further advance this plateau. By adding synthetic tumors---three times the number of real tumors---we achieved similar or better performance with only \lowerbound\ real tumor scans. This reduces annotation needs by a large margin and shows that synthetic data can accelerate learning, effectively steepening the scaling curve more than real data alone.

The lesson on the proprietary dataset helps estimate how many annotated tumor scans are needed to train effective AI models, e.g., matching radiologist performance. Considering that pancreatic tumors are especially hard to detect on CT, with 80\% detected only at late stages \cite{henrikson2019screening,li2025pants}, we hypothesize that \emph{if \upperbound\ real scans---or \lowerbound\ with synthetic data---are enough for pancreatic tumors, the same or fewer might work for other organs.} Based on this idea, our \ul{first contribution} is to create a dataset, that is publicly available, with \lowerbound–\upperbound\ per-voxel annotated CT scans for tumors in six organs: pancreas, liver, kidney, colon, esophagus, and uterus. This is also the \textit{first} public dataset that offers per-voxel annotations for esophageal and uterine tumors. We name this six-tumor dataset \textbf{\ourdataset}, which comprises 4,242 CT scans with per-voxel annotations of \numoftumorinst\ benign/malignant tumor instances and \numofcontrolscans\ normal scans as control (\S\ref{sec:dataset}). Importantly, it includes many early-stage tumors ($<$20 mm): 5,709 in liver, 850 in pancreas, 4,638 in kidney, 29 in colon, 17 in esophagus, and 39 in uterus---rare and hard to collect.

While \ourdataset\ is much larger than public tumor datasets combined~\cite{heller2019kits19,ma2024fast,bilic2019liver,kang2023label}, the \lowerbound--\upperbound\ scans per tumor type are still insufficient for building robust AI across diverse data sources. This limitation is clear in our data-scaling analysis (\figureautorefname~\ref{fig:teaser}), where performance plateaued only on in-distribution tests. For out-of-distribution data---CT scans from different centers---performance kept improving up to \fullnumber\ scans, suggesting that broader diversity is critical for generalization. However, scaling to that level is costly: annotating just \lowerbound--\upperbound\ scans per tumor type required \numofradiologists\ radiologists and several months of effort. Selecting the most valuable scans to annotate is also challenging, since out-of-distribution data are unknown in advance.

\begin{table*}[h]
    \centering
    \begin{threeparttable}
    \scriptsize
    
    \begin{tabular}{p{0.155\linewidth}P{0.03\linewidth}P{0.065\linewidth}P{0.065\linewidth}P{0.06\linewidth}P{0.08\linewidth}P{0.07\linewidth}P{0.175\linewidth}P{0.08\linewidth}}
    \toprule
    Dataset & release  &\# scans &\# slices (K) & \#  tumors &  tumor in \# &\#  hospitals & \#  countries\textsuperscript{\ddag} & annotators    \\
    \midrule
    \rowcolor{iccvblue!10}LiTS~\cite{bilic2019liver}  \href{https://competitions.codalab.org/competitions/17094}{[link]} & 2019 & 131 & 58.6 & 853 & liver & 7 & E, NL, CA, FR, IL & human \\
    \rowcolor{white}MSD-Colon~\cite{antonelli2021medical}  \href{https://decathlon-10.grand-challenge.org/}{[link]} & 2021 & 126 & 13.5 &131 & colon & 1 & US & human \& AI\\
    \rowcolor{iccvblue!10}MSD-Pancreas~\cite{antonelli2021medical}  \href{https://decathlon-10.grand-challenge.org/}{[link]}& 2021 & 281 & 26.7 & 283 & pancreas & 1  & US & human \& AI \\
    \rowcolor{white}FLARE23 ~\cite{antonelli2021medical} \href{https://codalab.lisn.upsaclay.fr/competitions/12239}{[link]} & 2022 & 2,200 & 629.1 & 1,511 & unknown\textsuperscript{\dag} & 30 & N/A & human \& AI \\
    \rowcolor{iccvblue!10}KiTS~\cite{heller2023kits21}   \href{https://kits-challenge.org/kits23/}{[link]} &2023 & 489 & 250.9 & 568 & kidney & 1 & US & human \\
    \rowcolor{white}ULS-Liver~\cite{de2024uls23}  \href{https://uls23.grand-challenge.org/}{[link]} &2023 & 49 &6.3 & 49& liver & 1 &- & human \\
    \rowcolor{iccvblue!10}ULS-Pancreas~\cite{de2024uls23}  \href{https://uls23.grand-challenge.org/}{[link]} & 2023& 120 & 15.4 & 120 & pancreas & 1 & NI & human  \\
    \rowcolor{white}ULS-Kidney~\cite{de2024uls23}  \href{https://uls23.grand-challenge.org/} {[link]} &2023 &50 & 6.4 &50 & kidney & 1 & N/A & human \\
    \midrule
    \rowcolor{iccvblue!10}\ourdataset\ (ours) &2025 &  \numofscans\ & 4,700 & \numoftumorinst & \makecell{liver, \\pancreas, \\kidneys, \\colon, \\esophagus,\\uterus} &  89 & \makecell{MT, IE, BR, BA, AUS,\\ TH, CA, TR, CL, ES, \\MA, US, DE, NL, FR, IL, CN} & human\\
    \bottomrule
    \end{tabular}
    \begin{tablenotes}
    \item[\dag]Tumors labeled in the FLARE23 dataset fall under a general 'Tumor' category without specific tumor type information.
    \item[\ddag ]US: United States, DE: Germany, NL: Netherlands, CA: Canada, FR: France, IL: Israel, IE: Ireland, BR: Brazil, BA: Bosnia and Herzegowina, CN: China, TR:
    Turkey, CH: Switzerland, AUS: Australia, TH: Thailand, CL: Chile, ES: Spain, MA: Morocco, and MT: Malta. 
    \end{tablenotes}
\end{threeparttable}
\caption{\textbf{Dataset comparison.} 
We compare \ourdataset\ against existing abdominal tumor segmentation datasets, including those with and without tumor labels.  \ourdataset\ outperforms these datasets in terms of scale and diversity.
}
   \label{tab:annotated_tumor_datasets}
\end{table*}

To address this, our \ul{second contribution} is to scale data and annotations through DiffTumor to produce different types of tumors (\figureautorefname~\ref{fig:synthetic_data_analysis}). The data-scaling analysis (\figureautorefname~\ref{fig:teaser}) suggested that training AI on synthetic tumors can significantly enhances in-distribution test performance. More importantly, since collecting normal scans is much easier than acquiring and annotating tumor scans, synthetic tumors can be added to normal scans from a range of out-of-distribution sources, bypassing the need for manual per-voxel annotation. These synthetic tumors are automatically paired with per-voxel annotations as they are generated with their masks. Training AI on these normal scans augmented by synthetic tumors can greatly improve performance in out-of-distribution tests (\figureautorefname~\ref{fig:fig_scaling_analysis_for_generalizability}).

In summary, we bring data-scaling lessons from both real and synthetic data on a large proprietary dataset to develop \ourdataset, achieving two key advancements for six-tumor segmentation, specifically,

\begin{enumerate}
    \item Scaling real and synthetic data enhances performance in abdominal tumor segmentation.  We rank first in the MSD challenge, leading to substantial performance improvement. We also achieve the highest performance on the validation sets of our \ourdataset\ dataset, improving DSC scores by +5\%, +9\%, +3\%, +4\%, +7\%, and +2\% for segmenting tumors in the liver, pancreas, kidney, colon, esophagus, and uterus, respectively, compared to the runner-up algorithms
    (\S\ref{sec:advantages_cancerverse}, Tables~\ref{tab:msd_test}--\ref{tab:benchmark_on_validation}).
    
    \item Scaling real and synthetic data enhances generalizable performance in abdominal tumor segmentation without additional tuning and adaptation. \ourdataset\ significantly outperforms the runner-up algorithms by +14\% DSC on four external datasets (\S\ref{sec:advantages_cancerverse}, \tableautorefname~\ref{tab:advantage_for_generalizability}).
\end{enumerate}

\section{Related Work}
\label{sec:related_work}

\smallskip\noindent\textbf{Large-scale Annotated Tumor Datasets} are scarce due to the limited availability of scan data and the substantial costs of obtaining per-voxel annotations. Despite these hurdles, datasets such as DeepLesion~\cite{yan2017deeplesion}, AutoPET~\cite{gatidis2024results}, PANORAMA~\cite{alves2024panorama}, FLARE~\cite{ma2024fast}, and MSD~\cite{antonelli2022medical} serve as significant efforts to mitigate this limitation. A detailed comparison of related datasets is provided in \figureautorefname~\ref{tab:annotated_tumor_datasets}.
\ourdataset\ comprises more than 10,000 CT scans with voxel-level annotations across six abdominal tumors. Notably, \ourdataset\ features esophageal and uterine tumor scans, which have not been previously available in public datasets.

\smallskip\noindent\textbf{Neural Scaling Laws} establish the power-law relationships that correlate model performance with key scaling factors such as model size, dataset volume, and computational resources. It is initially discovered within the domain of language models highlighted by Kaplan \etal \cite{kaplan2020scaling}, and soon also been observed in generative visual modeling \cite{peebles2023scalable, henighan2020scaling} and multi-modality modeling \cite{jia2021scaling}. This trend of scaling underpins the recent achievements of foundation models \cite{ouyang2022training, rombach2022high}, emphasizing how scaling up systematically boosts model generalization and effectiveness across various tasks. However, for tumor analysis and synthetic data, scaling laws remain underexplored due to the limited availability of annotated tumor data. Leveraging our new, large-scale tumor dataset, we investigate whether similar data scaling laws exist in tumor segmentation and whether appropriate data scaling can yield a robust segmentation model capable of generalizing to detect and segment tumors from CT scans, encompassing a broad spectrum of patient demographics, imaging protocols, and healthcare facilities.

\section{\ourdataset}
\label{sec:dataset}

\subsection{Dataset Construction}
\label{sec:construction_dataset}

\begin{figure*}[t]
	\centering
	\includegraphics[width=\linewidth]{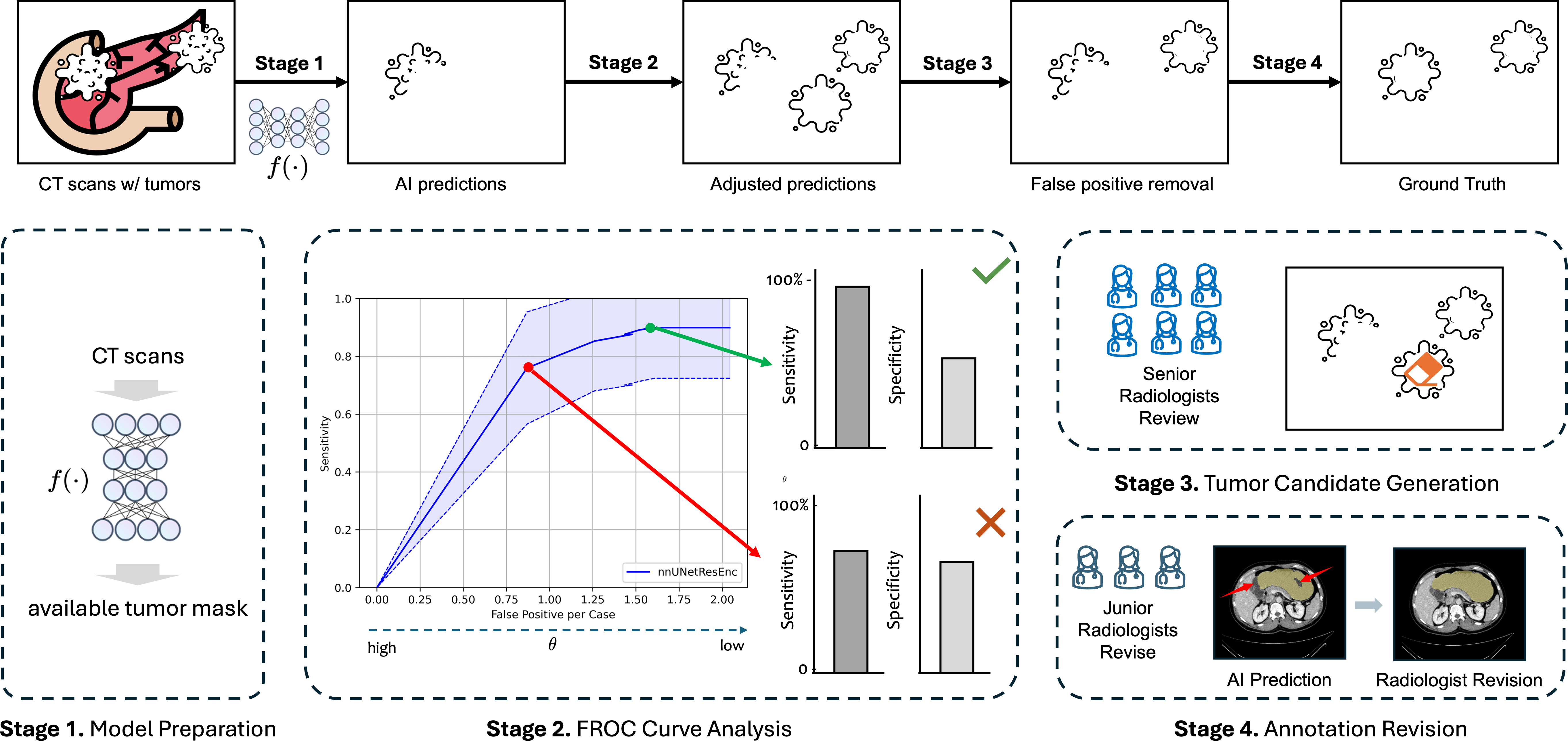}
	\caption{\textbf{ Overview of the \ourtool.} Towards annotating a large-scale tumor dataset, developing our \ourtool\ involves four stages. \ding{172} Train a Segmentation Model using public datasets to provide tumor segmentation logits across \ourdataset. \ding{173} Analyzing the FROC curve and selecting a threshold that enhances sensitivity to minimize missed tumors while maintaining an acceptable specificity score. \ding{174} Removing false positives for the adjusted predictions by senior radiologists.  \ding{175} Revising the final annotations to get ground truth by junior radiologists. }
    \label{fig:efficient_tumor_annotation}
\end{figure*}

Accurate annotations are the foundation of high-quality medical datasets. However, conventional per-voxel labeling is labor-intensive. Obtaining each scan data typically costs 4--5 minutes, while extensive tumors may take up to 40 minutes~\cite{bilic2023liver,ma2024fast}. In addition, precisely delineating tumor boundaries takes substantial time and requires the specialized expertise of highly trained radiologists, making it impractical to scale annotations to datasets with 10,000 or more scans.  To address this bottleneck, we establish a semi-automated annotation pipeline for CT scans that significantly reduces the manual workload and requires only minimal revision time from radiologists.  

\smallskip\noindent\textbf{SMART-Annotator Procedure.} Annotating missed tumors from scratch takes much longer than removing AI-generated false positives. Therefore, our annotation pipeline is designed to prioritize minimizing under-segmentation errors, thereby reducing the typical annotation time from 5 minutes per scan to less than 5 seconds on average, while maintaining high accuracy.  The proposed pipeline, named \ourtool, stands for \textit{Segmentation Model-Assisted Rapid Tumor Annotator}. As depicted in \figureautorefname~\ref{fig:efficient_tumor_annotation}, it consists of the following four key stages:

\smallskip\noindent\textbf{Stage 1: Model Preparation.} For each tumor, we separately train a Segmentation Model (denoted as $f(\cdot)$) using publicly available datasets. The tumor-specific $f(\cdot)$ is optimized for tumor segmentation and detection tasks.
    
\smallskip\noindent\textbf{Stage 2: FROC Curve Analysis.}  To determine the optimal threshold, we construct the Free-response ROC  (FROC) Curve by equipping $f(\cdot, \theta)$ with a set of threshold values $\theta$, obtaining the trade-off map (as shown by the purple shadow region in \figureautorefname~\ref{fig:efficient_tumor_annotation}) between sensitivity and false positive rate. Experimental results on tumor analysis in CT scans reveal that a lower  $\theta^*$ maximizes sensitivity while maintaining an acceptable false positive rate.
 
\smallskip\noindent\textbf{Stage 3: Tumor Candidate Generation.}  For CT scans requiring annotation, we apply the tumor-specific model $f(\cdot, \theta^*)$ to perform voxel-wise analysis. This process generates preliminary tumor segmentation candidates, while identifying potential tumor regions that need further refinement and validation. Since these potential regions are typically challenging, senior radiologists are then required to conduct a review to confirm true positives and eliminate false positive cases.
    
\smallskip\noindent\textbf{Stage 4: Annotation Revision.}  The reviewed tumor segmentation candidates undergo further refinement by junior radiologists, who annotate missed tumors and adjust mask boundaries to ensure accurate and precise tumor annotations. The final revised annotations are thoroughly reviewed by senior radiologists to guarantee high-quality ground truth.

\smallskip\noindent\textbf{Annotation Accuracy Analysis.} For each specific organ, our pipeline adaptively adjusts the threshold $\theta^*$ based on the FROC curve to ensure over 90\% sensitivity. A common concern is whether such high sensitivity might result in a significant number of false positive cases? To answer this, we validate \ourtool\ on three public datasets and reveal that the pipeline maintains manageable false-positive rates,  with an average of 1.2 false positives per scan for pancreatic tumors, 2 for liver tumors, and 2.4 for kidney tumors.  These results highlight the effectiveness of our AI-driven approach in tumor detection. By pre-identifying tumors with pseudo-annotations, radiologists can quickly verify true positives, correct false positives, and, if necessary, provide additional annotations for false negatives, thereby efficiently annotating tumor scans in \ourdataset.

\begin{figure}[t]
	\centering
	\includegraphics[width=\linewidth]{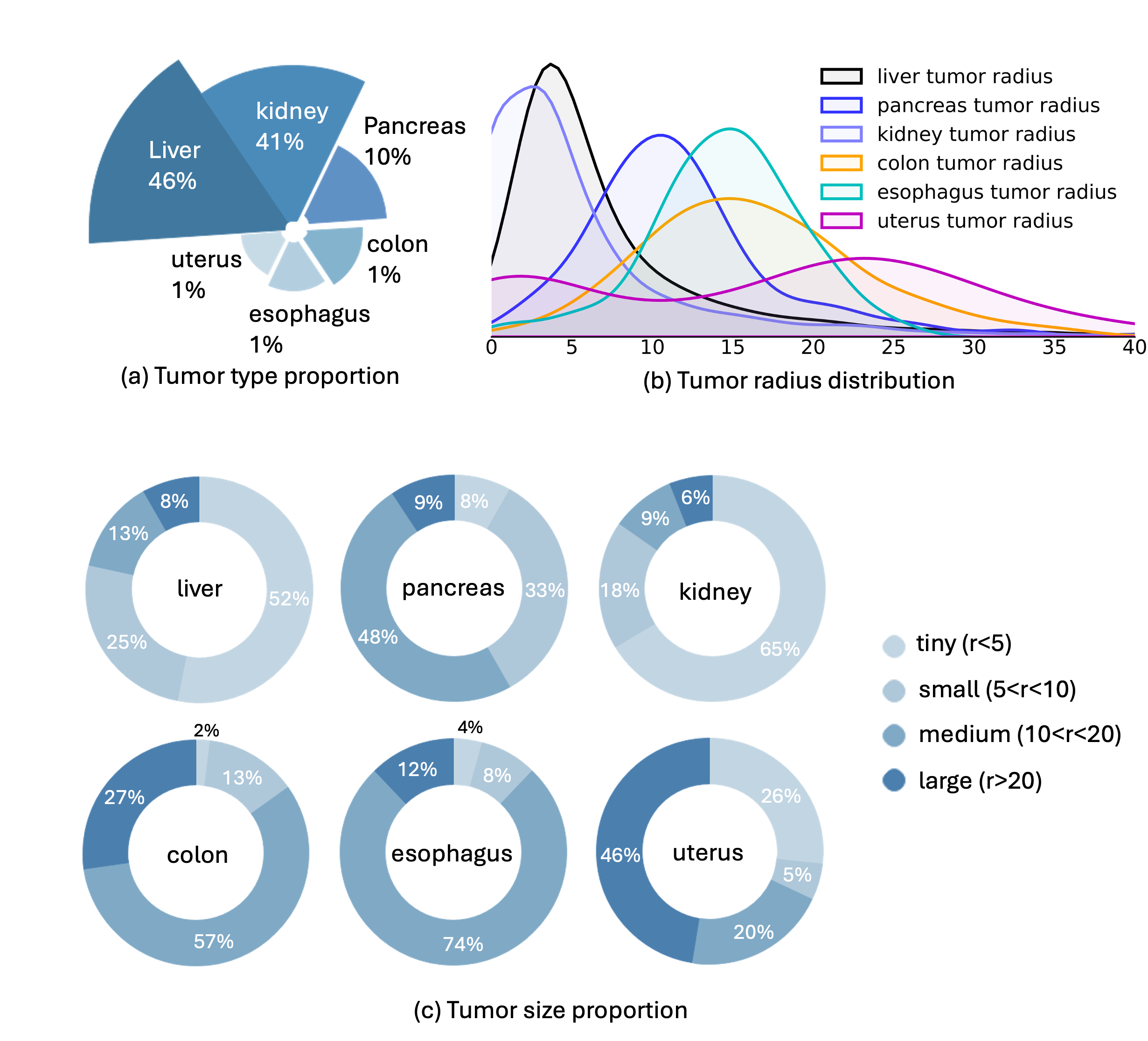}
	\caption{\textbf{Dataset statistics analysis} on the distributions of (a) different tumor proportions, (b) tumor radius, and (c) different tumor sizes categorized as tiny, small, medium, and large.
    }
	\label{fig:dataset_statistics}
\end{figure}

\smallskip\noindent\textbf{Annotation Efficiency Analysis.} The \ourdataset\ incorporates proprietary esophagus and uterus scans alongside unannotated data from 12 publicly available sources. Our approach applies the \ourtool\ pipeline to all scans. Given that full manual annotation typically requires 5 minutes per scan, whereas annotation with \ourtool\ takes only 5 seconds, this AI-driven approach substantially alleviates the annotation workload, conserving approximately \numofscans\ $\times \left(5 - \frac{1}{12}\right) \approx 49,826$ minutes of valuable radiologist time for annotating the entire \ourdataset\ collection. Assuming a radiologist works 10 hours per day, this corresponds to 83 workdays saved.

\subsection{Dataset Statistical Analysis}
\label{sec:advantages_dataset}

\ourdataset\ is the largest public, human-annotated tumor segmentation dataset, covering six tumor types (\figureautorefname~\ref{fig:ourdataset}). It improves on existing datasets in five ways:

\smallskip\noindent\textbf{1.~Large-Scale CT Coverage}. \ourdataset\ includes 10,136 fully annotated CT scans, totaling over 4.7 million slices. It provides labels for 31 anatomical structures, including 25 organs and 6 tumor types. The data comes from 89 hospitals, ensuring diverse patient populations and clinical conditions.
    
\smallskip\noindent\textbf{2.~Diverse Tumor Types}. Most public datasets focus on a single tumor type (\tableautorefname~\ref{tab:annotated_tumor_datasets}). In contrast, \ourdataset\ includes liver, pancreas, kidney, colon, esophageal, and uterine tumors. It is the first public dataset with voxel-wise annotations for esophageal and uterine tumors, supporting research on rare and underrepresented cancers (\figureautorefname~\ref{fig:dataset_statistics}a).

\smallskip\noindent\textbf{3.~Wide Tumor Size Range}. Tumor sizes in \ourdataset\ range from \textit{0} to \textit{100} mm. We group them into four categories: tiny ($r \leq 5$ mm), small ($5 \leq r \leq 10$ mm), medium ($10 \leq r \leq 20$ mm), and large ($r \geq 20$ mm). \ourdataset\ provides a balanced size distribution across all tumor types, as shown in \figureautorefname~\ref{fig:dataset_statistics}b--c), enabling robust and scalable model training.

\smallskip\noindent\textbf{4.~Abundant Tumor Masks}. The dataset contains 10,260 annotated tumor masks across six tumor types and all size groups. It surpasses existing datasets such as LiTS and KiTS in both scale and tumor diversity (\figureautorefname~\ref{fig:ourdataset}b, \tableautorefname~\ref{tab:annotated_tumor_datasets}).

\smallskip\noindent\textbf{5. High-Quality Annotations}. Our annotation pipeline adopts a multi-stage review process (see \figureautorefname~\ref{fig:efficient_tumor_annotation}), integrating AI algorithms with human expertise to enhance efficiency while maintaining high annotation quality.  All images and annotations undergo rigorous quality control. This process iteratively refined the annotations until no further major revisions were necessary.

\subsection{Advantages of \ourdataset}
\label{sec:advantages_cancerverse}
\noindent\textbf{Strong performance on in-distribution data.}
We report detailed comparisons on the official test set of the Medical Segmentation Decathlon (MSD) leaderboard in \tableautorefname~\ref{tab:msd_test}. As can be seen, with \ourdataset, we significantly surpass the previously leading Universal Model~\cite{liu2023clip} (denoted as Uni. Model in \tableautorefname~\ref{tab:msd_test}) and achieve the top \#1 performance on the leaderboard,  underscoring the superiority of \ourdataset\ in the task of medical segmentation. 

\begin{table}[t]
\centering
\scriptsize
\begin{tabular}
{p{0.26\linewidth}P{0.12\linewidth}P{0.12\linewidth}P{0.12\linewidth}P{0.12\linewidth}}
\toprule
 & \multicolumn{2}{c}{Task03 Liver} & \multicolumn{2}{c}{Task07 Pancreas}  \\
\cmidrule(lr){2-3}\cmidrule(lr){4-5}
Method & DSC & NSD & DSC & NSD \\ 
\midrule
Kim~\etal~\cite{kim2019scalable} & 73.0  & 88.6      & 51.8  & 73.1   \\
C2FNAS~\cite{yu2020c2fnas} & 72.9  & 89.2    & 54.4   & 75.6  \\
Trans VW~\cite{haghighi2021transferable}  & 76.9\scalebox{.6}{$\pm$20.0} & {92.0\scalebox{.6}{$\pm$16.8}} & 51.1\scalebox{.6}{$\pm$32.8} & 70.1\scalebox{.6}{$\pm$37.4} \\
Models Gen.~\cite{zhou2021models} & {77.5\scalebox{.6}{$\pm$20.4}} & 91.9\scalebox{.6}{$\pm$17.9} & 50.4\scalebox{.6}{$\pm$32.6} & 70.0\scalebox{.6}{$\pm$37.2} \\
nnU-Net~\cite{isensee2021nnu} & 76.0\scalebox{.6}{$\pm$22.1} & 90.7\scalebox{.6}{$\pm$18.3} & 52.8\scalebox{.6}{$\pm$33.0} & 71.5\scalebox{.6}{$\pm$36.6} \\
DiNTS~\cite{he2021dints} & 74.6\scalebox{.6}{$\pm$21.3} & 91.0\scalebox{.6}{$\pm$17.3} & 55.4\scalebox{.6}{$\pm$29.8} & 75.9\scalebox{.6}{$\pm$32.0} \\
Swin UNETR~\cite{tang2022self} & 75.7\scalebox{.6}{$\pm$20.4} & 91.6\scalebox{.6}{$\pm$16.8} & {58.2\scalebox{.6}{$\pm$28.6}} & {79.1\scalebox{.6}{$\pm$29.7}} \\
Uni. Model~\cite{liu2023clip} & 79.4\scalebox{.6}{$\pm$17.0} & 93.4\scalebox{.6}{$\pm$15.2} & 62.3\scalebox{.6}{$\pm$26.6} & 82.9\scalebox{.6}{$\pm$27.2} \\
\midrule
\ourdataset & \textbf{82.6\scalebox{.6}{$\pm$11.0}} & \textbf{96.9\scalebox{.6}{$\pm$6.4}} & \textbf{67.2\scalebox{.6}{$\pm$24.7}} & \textbf{86.0\scalebox{.6}{$\pm$25.2}} \\ 
$\Delta$ & \textbf{\textcolor{red}{+3.2}} & \textbf{\textcolor{red}{+3.5}} & \textbf{\textcolor{red}{+4.9}} & \textbf{\textcolor{red}{+3.1}} \\
\bottomrule
\end{tabular}
\caption{\textbf{Leaderboard performance on MSD Challenge.} 
The results are assessed on the MSD official server using the MSD competition test dataset. All DSC and NSD metrics are sourced from \href{https://decathlon-10.grand-challenge.org/evaluation/challenge/leaderboard/}{The MSD Leaderboard}. The outcomes for the remaining tasks were produced by Universal Model~\cite{liu2023clip,liu2024universal}.}
\label{tab:msd_test}
\end{table}

To comprehensively evaluate the six tumor types in \ourdataset, we 
train ResEncM~\cite{isensee2024nnu} with the annotated tumor data in \ourdataset\ and compare with state-of-the-art segmentation models in the medical field (\ie, UNETR~\cite{hatamizadeh2022unetr}, Swin UNETR~\cite{tang2022self}, nnU-Net~\cite{isensee2021nnu}, ResEncM~\cite{isensee2024nnu} and STU-Net-B~\cite{huang2023stu}) that are trained with publicly available tumor datasets. The evaluations are conducted on the validation set of \ourdataset\  and reported in Table~\ref{tab:benchmark_on_validation}. As can be seen,  training the ResEncM with \ourdataset\ (denoted as \ourdataset) consistently improves the performance and outperforms the state-of-the-art across all tumor segmentation tasks. Compared with the second-ranked STU-Net-B,  \ourdataset\ archives a remarkable DSC improvement of \textit{7.3\%} on esophageal tumors and \textit{4.9\%} on liver tumors, respectively.
These results demonstrate the superiority of \ourdataset\ in delivering high-quality tumor data for model training compared to existing datasets, contributing to alleviating the data scarcity issue in tumor segmentation.

\begin{table*}[t]
    \centering
    \scriptsize
    \begin{tabular}{p{0.14\linewidth}P{0.08\linewidth}P{0.09\linewidth}P{0.04\linewidth}P{0.04\linewidth}P{0.09\linewidth}P{0.04\linewidth}P{0.04\linewidth}P{0.09\linewidth}P{0.04\linewidth}P{0.04\linewidth}} 
    \toprule
     & & \multicolumn{3}{c}{Liver Tumor} & \multicolumn{3}{c}{Pancreatic Tumor} & \multicolumn{3}{c}{Kidney Tumor} \\
    \cmidrule(lr){3-5}\cmidrule(lr){6-8}\cmidrule(lr){9-11}
    Method & Param & Sen. & DSC & NSD & Sen. & DSC & NSD
    & Sen. & DSC & NSD \\
    \midrule
    UNETR~\cite{hatamizadeh2022unetr} &101.8M &77.1\tiny\textcolor{gray}{~(102/131)} &55.6 &53.7 &66.7\tiny\textcolor{gray}{~(102/131)} & 31.1 &27.2 &95.8\tiny\textcolor{gray}{~(102/131)} &67.2&55.7\\
    Swin UNETR~\cite{tang2022self} &72.8M &76.6\tiny\textcolor{gray}{~(102/131)} &66.8&68.4&81.5\tiny\textcolor{gray}{~(102/131)} & 44.7 &43.8 &95.8\tiny\textcolor{gray}{~(102/131)} &72.3 &67.7 \\
    nnU-Net~\cite{isensee2021nnu} &31.1M &80.3\tiny\textcolor{gray}{~(102/131)} &71.7 &74.6 &81.5\tiny\textcolor{gray}{~(102/131)} &56.7 &54.3 &100\tiny\textcolor{gray}{~(102/131)} & 84.8 &80.7\\
    ResEncM~\cite{isensee2024nnu}&63.1M   &\textbf{89.1}\tiny\textcolor{gray}{~(102/131)} &71.9 &74.7 &84.0\tiny\textcolor{gray}{~(102/131)} &57.0 &54.6 &100\tiny\textcolor{gray}{~(102/131)} &84.8 &81.1\\
    STU-Net-B~\cite{huang2023stu}&58.3M &79.3\tiny\textcolor{gray}{~(102/131)} &72.6 &74.9 &85.2\tiny\textcolor{gray}{~(102/131)} &56.1&54.4 &100\tiny\textcolor{gray}{~(102/131)} &82.4 &77.6\\
    \midrule
    \ourdataset  &63.1M  &83.7\tiny\textcolor{gray}{~(102/131)}  &\textbf{77.5}&\textbf{81.0} &\textbf{96.0}\tiny\textcolor{gray}{~(102/131)} & \textbf{65.8} &\textbf{64.7} &\textbf{100}\tiny\textcolor{gray}{~(102/131)} &\textbf{87.9} &\textbf{84.4} \\
    $\Delta$ & & \textbf{\textcolor{blue}{-5.4}} &\textbf{\textcolor{red}{+4.9}} &\textbf{\textcolor{red}{+6.1}} &\textbf{\textcolor{red}{+10.8}} &\textbf{\textcolor{red}{+8.8}}  &\textbf{\textcolor{red}{+10.1}} & \textbf{\textcolor{red}{+0.0}} &\textbf{\textcolor{red}{+3.1}} &\textbf{\textcolor{red}{+3.3}} \\
    \midrule
     & & \multicolumn{3}{c}{Colon Tumor} & \multicolumn{3}{c}{Esophagus Tumor} & \multicolumn{3}{c}{Uterus Tumor} \\
    \cmidrule(lr){3-5}\cmidrule(lr){6-8}\cmidrule(lr){9-11}
    Method & Param & Sen. & DSC & NSD & Sen. & DSC & NSD
    & Sen. & DSC & NSD \\
    \midrule
    UNETR~\cite{hatamizadeh2022unetr} &101.8M &69.2\tiny\textcolor{gray}{~(102/131)} &27.8 &29.2 &92.3\tiny\textcolor{gray}{~(102/131)} &42.3 &44.1 &95.8\tiny\textcolor{gray}{~(102/131)} &69.9 &60.7\\
    Swin UNETR~\cite{tang2022self} &72.8M &65.4\tiny\textcolor{gray}{~(102/131)} &36.8 &39.4 &84.6\tiny\textcolor{gray}{~(102/131)} &48.2 &49.0 &95.8\tiny\textcolor{gray}{~(102/131)} &73.8 &65.0\\
    nnU-Net~\cite{isensee2021nnu}&31.3M &65.4\tiny\textcolor{gray}{~(102/131)} &42.8 &43.7 &92.3\tiny\textcolor{gray}{~(102/131)} &52.7 &53.2 &95.8\tiny\textcolor{gray}{~(102/131)} &78.5 &70.2\\
    ResEncM~\cite{isensee2024nnu} &63.1M &65.4\tiny\textcolor{gray}{~(102/131)} &43.8 &45.9 &84.6\tiny\textcolor{gray}{~(102/131)} &53.3 &51.9 &95.8\tiny\textcolor{gray}{~(102/131)} &78.7 &68.4\\
    STU-Net-B~\cite{huang2023stu} &58.3M &73.1\tiny\textcolor{gray}{~(102/131)} &47.1 &\textbf{48.7} &88.5\tiny\textcolor{gray}{~(102/131)} &53.9 &54.1 &95.8\tiny\textcolor{gray}{~(102/131)} &78.2 &68.8\\
    \midrule
    \ourdataset    &63.1M &\textbf{96.2}\tiny\textcolor{gray}{~(102/131)} &\textbf{50.7}  &47.6 &\textbf{96.2}\tiny\textcolor{gray}{~(102/131)} &\textbf{61.2} &\textbf{61.7}  & 95.8\tiny\textcolor{gray}{~(102/131)} &\textbf{80.1} &\textbf{70.3}\\
    $\Delta$ & & \textbf{\textcolor{red}{+23.1}} &\textbf{\textcolor{red}{+3.6}} & \textbf{\textcolor{blue}{-1.1}} &\textbf{\textcolor{red}{+3.9}} &\textbf{\textcolor{red}{+7.3}}  &\textbf{\textcolor{red}{+7.6}} & \textbf{\textcolor{red}{+0.0}} &\textbf{\textcolor{red}{+1.4}} &\textbf{\textcolor{red}{+0.1}} \\
    \bottomrule
    \end{tabular} 
    \caption{\textbf{Strong performance for in-distribution data: Results on \ourdataset}.
    We compare \ourdataset\ with common AI algorithms, using the validation sets from the \ourdataset. \ourdataset\ demonstrates superior tumor segmentation and performance overall, showing significant improvements in segmenting liver tumors (+4.9\%), pancreatic tumors (+8.8\%), kidney tumors (+3.1\%), colon tumors (+3.6\%), esophagus tumors (+7.3\%), and uterus tumors (+1.4\%). 
    } 
    \label{tab:benchmark_on_validation}
\end{table*}

\smallskip\noindent\textbf{Better generalization for out-of-distribution data.}
A critical requirement for medical AI models is their ability to generalize across diverse, out-of-distribution (OOD) data from multiple hospitals, rather than being optimized solely for a single, in-distribution dataset.  As shown in \tableautorefname~\ref{tab:annotated_tumor_datasets}, \ourdataset\ provides a considerably more diverse collection of CT scans from 89 hospitals across 18 countries. To verify the generalizability offered by \ourdataset, we further conduct evaluations on four external datasets: 3D-IRCADb~\cite{soler20103d}, PANORAMA~\cite{alves2024panorama}, Kipa~\cite{he2021meta}, and a proprietary JHH dataset \cite{xia2022felix}, none of which are included in the training phase. We train ResEncM~\cite{isensee2024nnu} with the annotated tumor data in \ourdataset\ and compare with the following state-of-the-art medical image segmentation models: UNETR~\cite{hatamizadeh2022unetr}, Swin UNETR~\cite{tang2022self}, nnU-Net~\cite{isensee2021nnu}, ResEncM~\cite{isensee2024nnu} and STU-Net~\cite{huang2023stu}, SegResNet~\cite{myronenko20193d}, Universal Model~\cite{liu2023clip}, and SuPrem~\cite{li2024abdomenatlas}. As shown in Table~\ref{tab:advantage_for_generalizability}, our model significantly outperforms previous methods on all external datasets, achieving a notable DSC improvement of \textit{14.0\%} and an NSD improvement of \textit{17.0\%} on the 3D-IRCADb dataset.

\begin{figure}[t]
	\centering
	\includegraphics[width=\columnwidth]{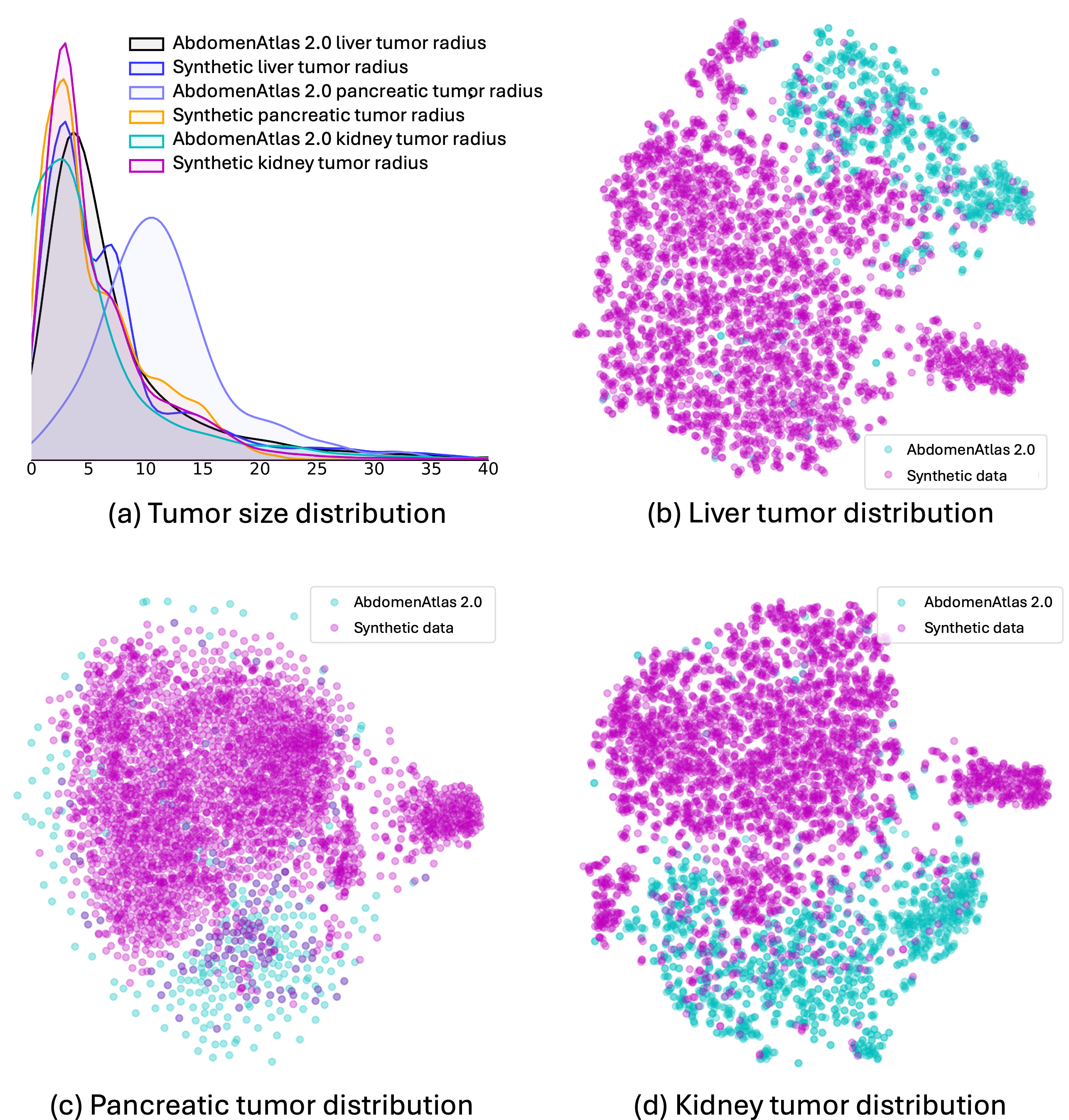}
    \caption{\textbf{Tumor size and feature distribution of real vs. synthetic tumors.} (a) Tumor size distribution across liver, pancreatic, and kidney tumors from real and synthetic data. (b–d) Feature distributions of liver, pancreatic, and kidney tumors. We extract features using a pretrained encoder~\cite{rombach2022high} and visualize them with t-SNE to compare synthetic tumors with real ones.
    } 
	\label{fig:synthetic_data_analysis}
\end{figure}

\begin{table*}[t]
    \centering
    \scriptsize
    \begin{tabular}{p{0.12\linewidth}P{0.07\linewidth}P{0.034\linewidth}P{0.034\linewidth}P{0.07\linewidth}P{0.034\linewidth}P{0.034\linewidth}P{0.07\linewidth}P{0.034\linewidth}P{0.034\linewidth}P{0.07\linewidth}P{0.034\linewidth}P{0.034\linewidth}} 
    \toprule
    &
    \multicolumn{3}{c}{3D-IRCADb~\cite{soler20103d} - Liver Tumor} & \multicolumn{3}{c}{PANORAMA~\cite{alves2024panorama} - Pancreatic Tumor} & \multicolumn{3}{c}{Kipa~\cite{he2021meta} - Kidney Tumor} & \multicolumn{3}{c}{JHH - Pancreatic Tumor}  \\
    \cmidrule(lr){2-4}\cmidrule(lr){5-7}\cmidrule(lr){8-10}\cmidrule(lr){11-13}
    Method & Sen. & DSC & NSD & Sen. & DSC & NSD
    & Sen. & DSC & NSD & Sen. & DSC & NSD
    \\
    \midrule
    UNETR~\cite{hatamizadeh2022unetr} &74.4\tiny\textcolor{gray}{~(87/117)} &50.1&46.8 &58.8\tiny\textcolor{gray}{~(77/131)} &21.4 &18.0 &70.8\tiny\textcolor{gray}{~(51/72)} &43.1 &35.8 & 51.4\tiny\textcolor{gray}{~(152/296)}& 13.0& 9.0\\
    Swin UNETR~\cite{tang2022self}  &76.9\tiny\textcolor{gray}{~(90/117)} &57.9&53.7 &69.5\tiny\textcolor{gray}{~(91/131)} &34.0 &30.9 &81.9\tiny\textcolor{gray}{~(59/72)} &64.3 &56.6 & 71.3\tiny\textcolor{gray}{~(211/296)}& 31.9& 21.9\\
    nnU-Net~\cite{isensee2021nnu} &77.8\tiny\textcolor{gray}{~(91/117)} &65.1 &62.2 &75.6\tiny\textcolor{gray}{~(99/131)} &42.4 &38.6 &80.6\tiny\textcolor{gray}{~(58/72)} &64.3 &58.9 & 69.9\tiny\textcolor{gray}{~(207/296)}& 34.1& 24.7 \\
    ResEncM~\cite{isensee2024nnu} &76.9\tiny\textcolor{gray}{~(90/117)} &57.6 &53.3 &61.1\tiny\textcolor{gray}{~(80/131)} &33.5 &30.0 &90.2\tiny\textcolor{gray}{~(65/72)} &76.4 &77.0 & 68.6\tiny\textcolor{gray}{~(203/296)}& 34.8& 26.5\\ 
    STU-Net~\cite{huang2023stu} &78.6\tiny\textcolor{gray}{~(92/117)} &67.1 &64.5 &74.0\tiny\textcolor{gray}{~(97/131)} &42.7 &40.3 &55.6\tiny\textcolor{gray}{~(40/72)} &71.2 &70.4 & 68.9\tiny\textcolor{gray}{~(204/296)}& 34.1& 24.7\\
    SegResNet~\cite{myronenko20193d}  & 65.0\tiny\textcolor{gray}{~(76/117)}  & 54.6 &51.3 &  84.0\tiny\textcolor{gray}{~(110/131)} & 43.0 & 40.3 &  94.4\tiny\textcolor{gray}{~(68/72)} & 73.6  & 70.0 & 77.7\tiny\textcolor{gray}{~(211/296)}& 39.5& 31.1 \\
    Universal Model~\cite{liu2023clip}   & \textbf{86.3}\tiny\textcolor{gray}{~(101/117)} & 62.8 & 57.4 & 77.9\tiny\textcolor{gray}{~(102/131)} & 37.0 & 33.9& \textbf{97.2}\tiny\textcolor{gray}{~(67/72)} & 47.8 & 37.1  & 78.4\tiny\textcolor{gray}{~(232/296)} & 32.6 & 27.1\\
    SuPreM~\cite{li2024abdomenatlas} & 58.1\tiny\textcolor{gray}{~(68/117)} & 50.2 & 47.8 & 67.9\tiny\textcolor{gray}{~(89/131)} & 30.5 & 28.0 & 84.7\tiny\textcolor{gray}{~(61/72)} & 42.3 & 36.0 & 63.2\tiny\textcolor{gray}{~(187/296)} & 24.7 & 19.8\\
    \midrule
    \ourdataset    & \textbf{86.3}\tiny\textcolor{gray}{~(101/117)} & \textbf{81.1} & \textbf{81.5}  &\textbf{94.6}\tiny\textcolor{gray}{~(124/131)} &\textbf{55.3} &\textbf{52.2} &\textbf{97.2}\tiny\textcolor{gray}{~(70/72)} &\textbf{83.6} &\textbf{83.0} &\textbf{80.7}\tiny\textcolor{gray}{~(239/296)} &\textbf{45.1} &\textbf{35.7}\\
    $\Delta$ & \textbf{\textcolor{red}{+0.0}} &\textbf{\textcolor{red}{+14.0}} &\textbf{\textcolor{red}{+17.0}} &\textbf{\textcolor{red}{+10.6}} &\textbf{\textcolor{red}{+12.3}} &\textbf{\textcolor{red}{+11.9}} & \textbf{\textcolor{red}{+0.0}} &\textbf{\textcolor{red}{+7.2}} & \textbf{\textcolor{red}{+6.0}} &\textbf{\textcolor{red}{+2.3}} &\textbf{\textcolor{red}{+5.6}} &\textbf{\textcolor{red}{+4.6}}\\
    \bottomrule
    \end{tabular}
    \caption{\textbf{Better generalizability for out-of-distribution data: Results on external datasets.} We evaluate \ourdataset\ and 8 other models on data from three publicly available and one private external source without additional fine-tuning or domain adaptation. Compared to dataset-specific models, \ourdataset\ demonstrates greater robustness when handling CT scans obtained from a variety of scanners, protocols, and institutes. 
    } 
    \label{tab:advantage_for_generalizability}
\end{table*}

\section{Scaling Laws in Tumor Segmentation}
\label{sec:result}

In this section, we explore the existence of data scaling laws in tumor segmentation and assess whether appropriate data scaling can yield a robust segmentation model.  This segmentation model should be generalizable to detect and segment tumors from CT scans, handling a broad spectrum of patient demographics, imaging protocols, and healthcare facilities. Specifically, we first examine the impact of increasing the number of annotated real-tumor scans on in-distribution performance. Then we analyze how the scale of annotated real-tumor data influences the model’s ability to generalize to out-of-distribution tumor data. 

\subsection{Experimental Setup}
We evaluate the scaling behavior with two data setups: (1) only real-tumor scans, and (2) a combination of both synthetic and real tumor scans.   Since small tumors are rare in public datasets but crucial for clinical applications, we employ DiffTumor~\cite{chen2024towards} to generate synthetic tumors, with a ratio of 4:2:1 for small, medium, and large tumors, respectively. The total number of synthetic tumor scans generated is three times of \ourdataset. The distribution of tumor size and combined data distribution are illustrated in \figureautorefname~\ref{fig:synthetic_data_analysis}, where we combine the generated tumors with different scales of \ourdataset\ training set to train the supervised ResEncM~\cite{isensee2024nnu}. The evaluation is conducted with segmentation metrics (\ie, DSC, NSD) and detection metrics (\ie, tumor-level and patient level sensitivity), using the validation set of \ourdataset\ and six external datasets (3D-IRCADb, ULS-Liver, ULS-Pancreas, PANORAMA, Kipa, and JHH dataset).

\begin{figure}[t]
	\centering
\includegraphics[width=1.0\linewidth]{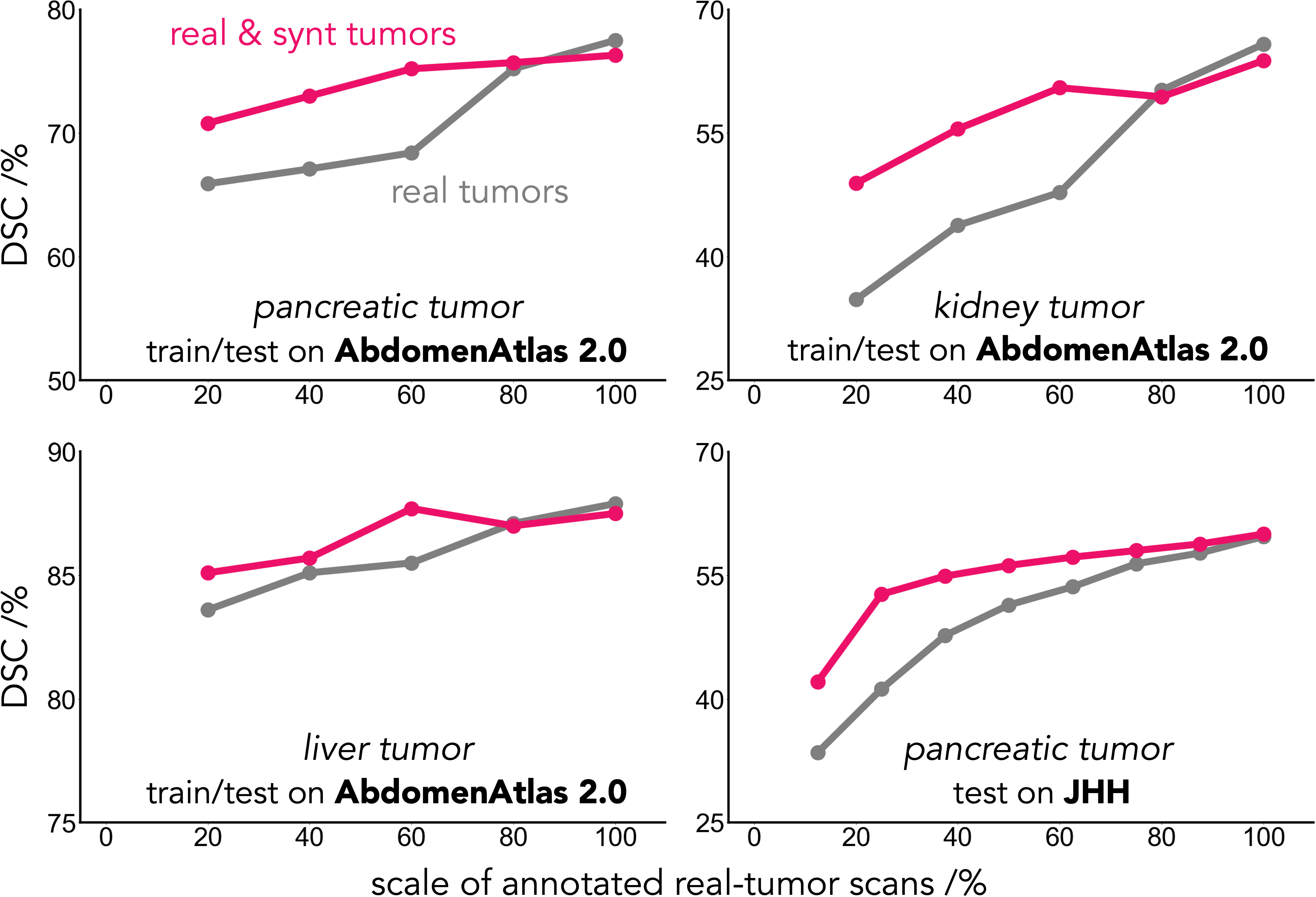}
	\caption{
    \textbf{Scaling data shows performance plateau in in-distribution evaluation.} We conduct a scaling study using \ourdataset\ and JHH datasets as real tumor data and evaluate performance on their corresponding validation sets. While scaling up the dataset initially enhances in-distribution performance, it eventually plateaus. These results align with the data-scaling lesson in \S\ref{sec:intro}. By supplementing real tumor data with well-designed synthetic data, we only need to collect and annotate a small amount of real data. This approach is especially beneficial for scenarios where data is scarce and annotation is costly, enabling high-accuracy segmentation with reduced effort.
    } 
	\label{fig:fig_scaling_analysis_for_iid}
\end{figure}

\begin{figure}[t]
	\centering
\includegraphics[width=\linewidth]{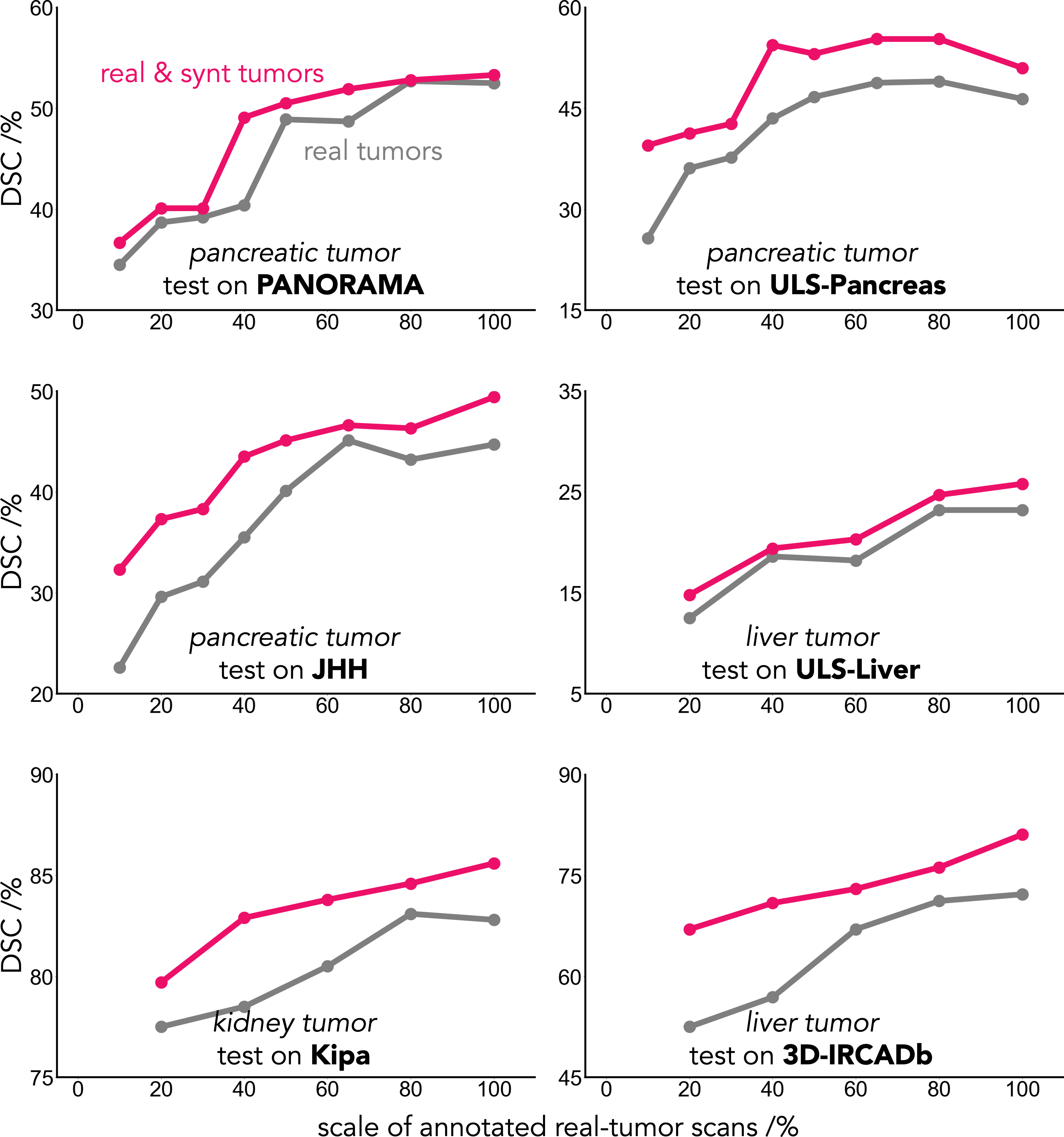}
	\caption{
    \textbf{Scaling data leads to greater generalizability.} 
    We conduct a scaling study using liver, pancreatic, and kidney tumors from the Cancerverse dataset as real tumor data and evaluate performance on six external datasets. Unlike in-distribution performance, which plateaus with more data, OOD generalization continues to improve with the addition of real tumor data. Notably, the integration of synthetic data further improves generalizability, with models trained on both real and synthetic scans consistently outperforming those using only real data. These findings underscore the critical importance of data diversity in enhancing model robustness across diverse imaging conditions. 
    }
	\label{fig:fig_scaling_analysis_for_generalizability}
\end{figure}

\subsection{Plateau in In-Distribution Evaluation}
\label{sec:scaling_analysis_with_real_data}
 We report the in-distribution segmentation performance in \figureautorefname~\ref{fig:fig_scaling_analysis_for_iid} and include the detection metrics in Appendix~\ref{sec:more_results_best_lesson_from_real_appendix}. Our analysis of tumor segmentation scaling behavior reveals a clear trend in in-distribution performance: as the number of annotated real-tumor scans increases, the in-distribution performance gains gradually saturate.  As illustrated by the gray lines in \figureautorefname~\ref{fig:fig_scaling_analysis_for_iid}, in-distribution performance initially improves with increasing data but eventually reaches a plateau across all three tumor types. This saturation indicates diminishing returns that adding more real tumor data yields progressively smaller performance gains. 

However, combining a certain amount of synthetic tumors with real data during training helps to accelerate this in-distribution performance saturation process.  As shown by the red lines in \figureautorefname~\ref{fig:fig_scaling_analysis_for_iid}, with the participants of synthetic data, the saturation status can be reached with only 40\% to 60\% of the annotated real-tumor scans, indicating that synthetic data effectively expedite the model’s convergence to its optimal performance within a given domain.  

This finding shows that we can achieve strong segmentation performance without collecting large amounts of real data. By supplementing real tumor data with well-designed synthetic data, we can significantly reduce the effort for costly real data annotation while maintaining strong in-distribution segmentation accuracy. This lesson demonstrates the tangible benefits of introducing synthetic data into the training process, and is particularly valuable for scenarios where real-data acquisition is costly or limited.

\subsection{Scaling Data Leads to Greater Generalizability} 
Figure \ref{fig:fig_scaling_analysis_for_generalizability} reports out‐of‐distribution (OOD) segmentation performance. As indicated by the gray and red lines, OOD accuracy consistently rises with the expanding dataset.  The impact of data scaling on out-of-distribution performance follows a consistently positive trend: as the amount of real tumor data increases, OOD performance continues to improve without signs of saturation, even after exhausting the entire Cancerverse dataset. We include more OOD results in Appendix~\ref{sec:more_results_best_lesson_from_synt_appendix}. In contrast with the in-distribution performance that tends to saturate with increasing data, the finding on the out-of-distribution performance reveals that OOD generalization continues to benefit from additional real tumor data without exhibiting diminishing returns. Such a non-diminishing trend is still obvious even when synthetic data is incorporated into the training process. Furthermore, models trained with both real and synthetic tumor scans consistently outperform those trained with real data.  
These findings underscore the critical role of data diversity in enhancing OOD generalization, showing that a carefully curated combination of real/synthetic data strengthens model robustness across diverse imaging settings.

\section{Conclusion}
\label{sec:conclusion}

This study examined not just whether more data helps tumor segmentation, but what kind of data is most valuable. Using \ourdataset---a large, expert-annotated dataset---and systematic scaling experiments, we learned three key lessons:
\ul{First}, performance on in-distribution data plateaus early. On internal datasets like JHH, segmentation accuracy stops improving after about 1,500 real scans. This means adding more similar data yields limited benefit once a moderate threshold is reached. \ul{Second}, synthetic tumors significantly reduce the need for manual annotation. By using synthetic lesions generated with DiffTumor, we can reach the same performance with only 500 real scans, cutting annotation effort by 70\%. Synthetic data improves data efficiency and accelerates model convergence. \ul{Third}, out-of-distribution generalization continues to improve with data diversity. Unlike the in-distribution case, performance on external datasets keeps increasing even after \upperbound\ scans and sees additional gains when synthetic tumors are added. This shows that model robustness depends more on data diversity than just quantity.

These lessons have important implications. Future expansion of \ourdataset\ should focus on including scans from different hospitals and imaging protocols. This will help the model perform better on new and unfamiliar data. For underrepresented tumor types like esophageal and uterine cancers, a few hundred well-selected scans combined with synthetic data can be enough to build useful models. \ourdataset\ also makes it possible to benchmark various data scaling and annotation strategies that were previously limited by small dataset size. The SMART-Annotator pipeline further shows how AI-assisted pre-labeling can reduce radiologist time from minutes to seconds per scan without sacrificing accuracy, especially when combined with synthetic tumor generation.

There are several limitations worth noting. First, the performance plateau observed at \upperbound\ scans applies only to pancreatic tumors in abdominal CT and with the ResEncM model. It is unclear whether this threshold holds for other organs, especially those with more complex or subtle tumor appearances. Future studies should examine how data scaling behaves across different tumor types, imaging modalities, and model architectures to see if similar saturation points occur. Second, although synthetic tumors improve performance, their anatomical realism---particularly for infiltrative, necrotic, or early-stage lesions---has not been fully verified by expert review or radiomic analysis. Ensuring clinically realistic synthesis remains a key challenge for building trust and interpretability.

\smallskip\noindent\textbf{Acknowledgments.}
This work was supported by the Lustgarten Foundation for Pancreatic Cancer Research and the National Institutes of Health (NIH) under Award Number R01EB037669. We would like to thank the Johns Hopkins Research IT team in \href{https://researchit.jhu.edu/}{IT@JH} for their support and infrastructure resources where some of these analyses were conducted; especially \href{https://researchit.jhu.edu/research-hpc/}{DISCOVERY HPC}.

{\small
\bibliographystyle{ieee_fullname}
\bibliography{refs,zzhou}
}

\newpage
\clearpage
\appendix
\begingroup
    \centering
    \Large
    \onecolumn
    \textbf{\thetitle}\\
    \vspace{0.5em}Supplementary Material
    \vspace{1em}\\
\endgroup

This appendix is organized as follows: 

\begin{itemize}
    \item \S\ \ref {sec:best_lesson_proof_appendix} provides comprehensive results with scaled real data with \jhhdataset\ dataset and synthetic data.
    \item  \S\ \ref {sec:related_works_appendix} provides comprehensive related works.
    \begin{itemize}
        \item[$\circ$] \ref{sec:AI_development_on_real_tumors_appendix}: AI Development on Real Tumors
        \item[$\circ$] \ref{sec:AI_development_on_synt_tumors_appendix}: AI Development on Synthetic Tumors 
    \end{itemize}
    \item  \S\ \ref {sec:implementation_details_appendix}  provides implementation details for \ourmodel\ and comparative models.
    \begin{itemize}
        \item[$\circ$]  \ref{sec:assembly_of_datasets_appendix}: details of public and private datasets used in \ourdataset.
        \item[$\circ$]  \ref{sec:comparative_models_appendix}: implementation details of comparative models
    \end{itemize}
    \item \S\ \ref {sec:visual_real_examples_appendix} provides more visual examples from \ourdataset.
    \item \S\ \ref{sec:more_results_best_lesson_from_real_appendix} presents additional results on the key insights gained from scaling real tumor data.
    \item \S\ \ref{sec:more_results_best_lesson_from_synt_appendix} presents additional results on the key insights gained from scaling real and synthetic tumor data.
\end{itemize}

\clearpage
\section{Best Lesson Proof on \jhhdataset\ dataset}
\label{sec:best_lesson_proof_appendix}

\begin{figure}[h]
	\centering
	\includegraphics[width=0.8\columnwidth]{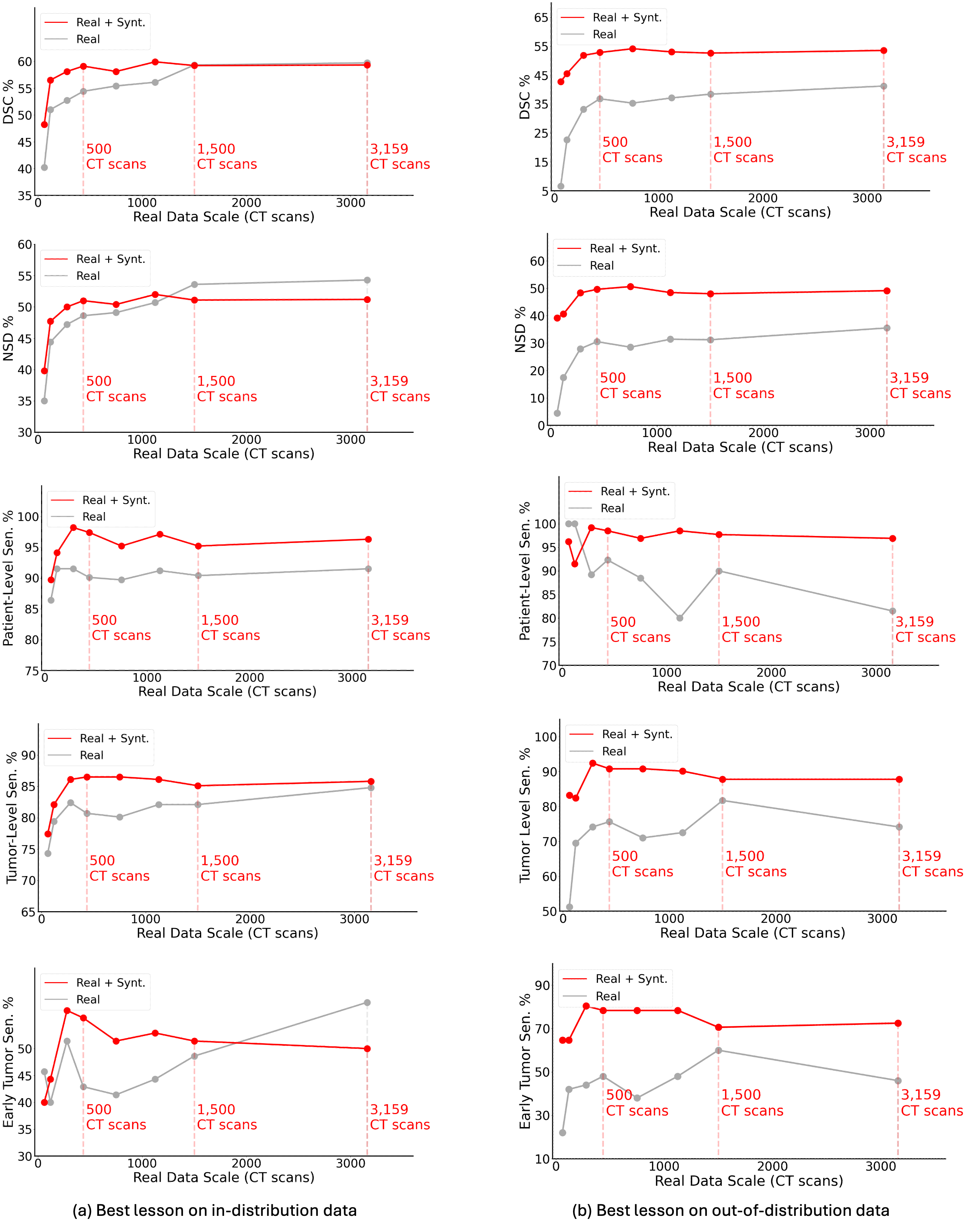}
    \caption{
    \textbf{Best lesson proof on \jhhdataset\ dataset.} Comprehensive experimental results trained on the \jhhdataset~dataset show that increasing the scale of real data (gray curve) improves segmentation (DSC and NSD) and detection (patient-level sensitivity, tumor-level sensitivity, and early tumor sensitivity)  for both in-distribution and out-of-distribution data. Additionally, augmenting the dataset with an extra $3\times$ synthetic data (red curve) consistently enhances the results. The specific numerical results in this figure can be referenced in the \tableautorefname~\ref{tab:supp_best_lesson_proof}. Given the substantial GPU requirements, the results were obtained from a single experiment. To reach a more reliable conclusion, we will conduct the experiments at least 10 times.
    } 
	\label{fig:supp_best_lesson_proof}
\end{figure}

\begin{table}[h]
\centering
\scriptsize
\begin{tabular}{p{0.2\linewidth} p{0.1\linewidth}P{0.1\linewidth}P{0.1\linewidth}P{0.06\linewidth}P{0.06\linewidth}P{0.12\linewidth}} 
\multicolumn{6}{l}{\textit{Scaling with real data.}} \\ \hline
&\#real CT & Patient-level Sen. & Tumor-level Sen. & DSC & NSD & Early Tumor Sen.\\
\shline
\multirow{8}{*}{Test on in-distribution data} & 60 & 86.4   & 74.3 & 40.2 &  35.0 & 45.7\\
& 120 & 90.1  &  79.4 &  51.0 & 44.4 & 40.0  \\
& 278 & 91.5  &  82.4 & 52.7 & 47.2 & 51.4 \\
& 435 & 90.1  & 80.7 &  54.4 & 48.6 & 42.9 \\
& 750 & 89.7  & 80.1 & 55.4 & 49.1 & 41.4 \\
& 1125 & 91.2  &  82.1 & 56.1 & 50.7 & 44.3 \\
& 1500 & 90.4  &  82.1 & 59.3 & 53.6 & 48.6 \\
& 3159 &91.5  & 84.8 & 59.7 & 54.3 & 58.6 \\ \hline
\multirow{8}{*}{Test on out-of-distribution data} & 60 & 100.0 & 51.2 &  6.6 & 4.4 &22.0\\
& 120 & 100.0  & 69.5& 22.6 & 17.4 & 42.0   \\
&  278 & 89.2  &  74.1 & 33.2 & 27.9 &44.0 \\
& 435 & 92.3  & 75.6 &  36.8 & 30.5 &48.0 \\
& 750 & 88.5 & 71.0 & 35.3 & 28.5 & 38.0 \\
& 1125 &  80.0  &  72.5 & 37.1 & 31.4 & 48.0 \\
& 1500 &  90.0  &  81.7 & 38.4 & 31.2 &  60.0\\
& 3159 & 81.5 & 74.1 & 41.2 & 35.5 & 46.0 \\
\hline
\end{tabular}

\begin{tabular}{p{0.2\linewidth} p{0.1\linewidth}P{0.1\linewidth}P{0.1\linewidth}P{0.06\linewidth}P{0.06\linewidth}P{0.12\linewidth}} \\
\multicolumn{6}{l}{\textit{Scaling with real \& synthetic data.}} \\ \hline
& \#real CT & Patient-level Sen. & Tumor-level Sen. & DSC & NSD & Early Tumor Sen.\\
\shline
\multirow{8}{*}{Test on in-distribution data}  & 60 &89.7  &77.4   &48.2  &39.8  &40.0  \\
&120 & 94.1  &82.1   &56.5  &47.7  &44.3  \\
&278 & 98.2    & 86.1  &58.1 &50.0  &57.1  \\
&435 & 97.4 &86.5   &59.1  &51.0  &55.7  \\
&750 &95.2     &86.5   &58.1  &50.4  &51.4  \\
&1125 &97.1     &86.1   &59.9  &52.0  &52.9  \\
&1500 &95.2     &85.1   &59.2  &51.1   &51.4  \\
&3159 &96.3   &85.8     &59.3   &51.2  &50.0  \\ \hline
\multirow{8}{*}{Test on out-of-distribution data} &60 &96.2     &83.2   &42.7  &39.2  &64.7  \\
&120 &91.5     & 82.4  &45.5  &40.6  & 64.7 \\
 &278 &99.2     & 92.4  &51.8  &48.3  &80.4  \\
&435 & 98.5    &90.8   &52.8  &49.6  &78.4  \\
&750 & 96.9    & 90.8  &54.1  &50.6  &78.4  \\
&1125 & 98.5    &90.1   &53.0  &48.4  &78.4  \\
&1500 &97.7     & 87.8  &52.6   &48.0  &70.6  \\
&3159 & 96.9    & 87.8  &53.5  &49.1  & 72.5 \\
\hline
\end{tabular}

\caption{\textbf{Best Lesson Proof on \jhhdataset\ dataset.} 
The \jhhdataset\ dataset comprises a total of 5,176 CT scans, which include scans of patients with pancreatic tumors as well as healthy scans without pancreatic tumors. We utilized 3,159 scans for training, while the remaining 2,017 were allocated for testing within the same distribution. For the out-of-distribution dataset, we selected the Panorama dataset. Detailed information regarding the dataset split can be found in \S\ \ref{sec:implementation_details_appendix}. For the segmentation model, we employed the SegResNet model based on the MONAI codebase for training and assessed the tumor segmentation and detection results using the DSC, NSD, and sensitivity metrics.}
\label{tab:supp_best_lesson_proof}
\end{table}

\clearpage
\section{Related works}
\label{sec:related_works_appendix}
\subsection{AI Development on Real Tumors}
\label{sec:AI_development_on_real_tumors_appendix}

\smallskip\noindent\textbf{\textit{AI algorithms.}}
Tumor detection and segmentation have been long-standing problems in medical image analysis. To achieve deliverable results, many recent works leverage state-of-the-art deep learning technology~\cite{tang2022self,huang2023stu}. 

The U-Net architecture~\cite{ronneberger2015u} has been widely adopted in medical image analysis. Over the years, numerous well-designed networks have been proposed to improve the U-Net architecture, including UNet++~\cite{zhou2018unet++,zhou2019unet++}, TransU-Net~\cite{chen2021transunet}, UNETR~\cite{hatamizadeh2022unetr}, Swin-UNETR~\cite{hatamizadeh2022swin}, and many others~\cite{cciccek20163d,milletari2016v,chen20233d,chen2021mt}. While these methods have demonstrated remarkable performance in tumor detection and segmentation, they typically rely on a significant number of annotations. The process of annotating real tumors is not only time-consuming but also requires extensive medical expertise. Sometimes, it needs the assistance of radiology reports \cite{bassi2025radgpt,bassi2025learning} or is even impossible to obtain the annotation~\cite{bilic2019liver,hu2023synthetic,xiang2024exploiting,yang2025medical}. Therefore, the use of synthetic tumors emerges as a promising solution.

 Liu et al.~\cite{liu2023clip} integrate text embeddings derived from Contrastive Language-Image Pre-training (CLIP) into segmentation models, effectively capturing anatomical relationships and enabling the model to learn structured feature embeddings across multiple organ and tumor types. With pre-training on large-scale CT scans with per-voxel annotations for 25 anatomical structures and seven tumor types, Li et al~\cite{li2024well} has developed a suite of models demonstrating robust transfer learning capabilities across various downstream organ and tumor segmentation tasks.

\smallskip\noindent\textbf{\textit{Preexisting public datasets}} have made significant contributions to the advancement of AI in tumor detection~\cite{li2025scalemai}. We summarizes key characteristics of existing public datasets for organ and tumor segmentation in table~\ref{tab:annotated_tumor_datasets}, categorized into those with and without tumor labels. Datasets such as LiTS~\cite{bilic2019liver} and KiTS~\cite{heller2023kits21} provide essential tumor labels but are limited with regard to  size and variety, with 131 and 489 scans, respectively, and fewer hospitals contributing data (7 for LiTS and 1 for KiTS). Larger datasets like FLARE23~\cite{liu2023flare} include 2,200 scans and span contributions from 30 hospitals, yet they focus on a single organ and provide no explicit tumor-specific labels. Similarly, datasets without tumor labels, such as WORD~\cite{luo2021word} and AMOS22~\cite{ji2022amos}, are useful for broader anatomical segmentation tasks but lack tumor-specific annotations. In contrast, \ourdataset\ distinguishes itself by offering the most extensive dataset to date, with 10,136 scans, 4,700K slices, and 13,223 tumors annotated across multiple organs, including rarer tumor types like esophagus and uterus. The dataset incorporates data from 89 hospitals across a wide range of countries, providing unprecedented diversity and comprehensiveness for multi-organ tumor research.

\subsection{AI Development on Synthetic Tumors}
\label{sec:AI_development_on_synt_tumors_appendix}
Tumor synthesis enables the generation of artificial tumors in medical images, aiding in the training of AI models for tumor detection and segmentation~\cite{jordon2018pate,yoon2019time,chen2021synthetic}. Synthetic tumors become particularly valuable when acquiring per-voxel annotations of real tumors is challenging, such as in the early stages of tumor development. There are several advantages of synthetic tumors over real tumors.

\smallskip\noindent\textbf{\textit{Quality Control}}: Synthetic data allows for the control of specific variables and the introduction of desired diversity into the dataset. Real-world datasets often suffer from imbalances, such as an overrepresentation of certain demographics or tumor stages. Synthetic data can be generated to balance these datasets, ensuring that machine learning models are trained on a comprehensive and representative sample of data. For rare cancers, collecting enough patient data is particularly difficult. Synthetic data can help augment these limited datasets, enabling the development of more robust and accurate models for rare cancer types. Additionally, synthetic data can be used to simulate hard cases that are difficult to capture in real-world data. Researchers can rapidly iterate and refine their models, leading to faster advancements in tumor detection, diagnosis, and treatment.

\smallskip\noindent\textbf{\textit{Privacy and Ethical Considerations}}: One of the major advantages of synthetic data is that it can be used without compromising patient privacy. Since synthetic data is not directly tied to any real individual, it eliminates the risk of exposing sensitive patient information. By using synthetic data, researchers can bypass ethical dilemmas associated with real patient data, such as the need for patient consent and the risk of data breaches.

Synthetic tumors can be used in aiding AI models for tumor detection and segmentation, particularly in situations where detailed annotations are scarce~\cite{domingos2012few, chen2021synthetic}. Therefore, an effective and universally applicable tumor synthesis approach is urgently needed to accelerate the development of tumor detection and segmentation methods.

\smallskip\noindent\textbf{\textit{Tumor development}} is intricately regulated by biological mechanisms at various scales. Tumors, which arise from DNA mutations in a single cell and represent genetic disorders~\cite{kumar2017robbins}, undergo complex growth processes. Mutated cells lead to uncontrolled proliferation, which can be benign or malignant~\cite{golias2004cell}. Differences between benign and malignant tumors include growth rate and invasiveness~\cite{kumar2017robbins}. Malignant tumors tend to exhibit larger final sizes and faster growth rates compared to benign lesions~\cite{kang2011cyst}. Additionally, slow tumor growth rates have been associated with low malignant potential~\cite{smaldone2012small,cheung2017active}. These patterns have also been observed in several studies~\cite{gui2018tumor,nathani2021hepatocellular}.
Malignant tumors usually invade surrounding tissues, while benign tumors typically remain confined to their original sites. Moreover, even slowly growing malignant tumors can invade surrounding tissues~\cite{kerr2016oxford}, leading to blurry boundaries between tumors and adjacent tissues. Therefore, it is necessary to design Accumulation and Growth rules to simulate these features. Tumor necrosis, a form of cell death, indicates a worse prognosis~\cite{richards2011prognostic,pollheimer2010tumor}. Histologically, necrosis is caused by hypoxia resulting from rapid cell proliferation surpassing vascular supply~\cite{hiraoka2010tumour}, presenting as non-enhancing irregular areas in CT images~\cite{fowler2021pathologic}. Hu et al.~\cite{hu2023label} developed a program that integrates medical knowledge to generate realistic liver tumors. However, these models are generally organ-specific and require adaptation to work with other organs. Lai et al.~\cite{lai2024pixel} proposed a framework that leverages cellular automata to simulate tumor growth, invasion, and necrosis, enabling realistic synthetic tumor generation across multiple organs.

\smallskip\noindent\textbf{\textit{Generative models}} 
have been effectively utilized in the medical field for tasks like image-to-image translation~\cite{lyu2022conversion, meng2022novel, ozbey2023unsupervised, pfeiffer2019generating}, reconstruction~\cite{song2021solving, xie2022measurement,lin2025pixel}, segmentation~\cite{fernandez2022can, kim2022diffusion, wolleb2022diffusion,chen2022mask}, and image denoising~\cite{gong2023pet}. Utilizing advanced generative models to synthesize various tumors is also a promising direction~\cite{guo2024maisi,wu2024freetumor,zhu2024generative,hamamci2025generatect}. Shin et al.~\cite{shin2018abnormal} advanced detection by generating synthetic abnormal colon polyps using Conditional Adversarial Networks.  Chen et al.~\cite{chen2024towards} employed a diffusion model that capitalizes on similarities in early-stage tumor imaging for cross-organ tumor synthesis. Wu et al.~\cite{wu2024freetumor} employs an adversarial-based discriminator to automatically filter out the low-quality synthetic tumors to  improve tumor synthesis. Guo et al.~\cite{guo2024maisi} incorporates ControlNet to process organ segmentation as additional conditions to guide the generation of CT images with flexible volume dimensions and voxel spacing.

\clearpage
\section{Implementation Details}
\label{sec:implementation_details_appendix}
\subsection{Dataset Composition}
\label{sec:assembly_of_datasets_appendix}

\begin{table*}[h]

\centering
\footnotesize
\begin{tabular}{p{0.02\linewidth}p{0.3\linewidth}P{0.09\linewidth}P{0.35\linewidth}P{0.085\linewidth}
}
\toprule
\multicolumn{2}{l}{\makecell[tl]{\ourdataset\ components}} & \makecell{\# of scans } & \makecell{annotated tumor (original)} & \makecell{annotators} \\
\midrule
\rowcolor{blue!10}\multicolumn{2}{l}{Public CT in \ourdataset\ (AbdomenAtlas1.1)}
& 9,901 & liver, pancreas, kidney, colon  & human \& AI \\
 & CHAOS \citeyearpar{valindria2018multi} [\href{https://chaos.grand-challenge.org/Download/}{link}] & 20  & - & human \\
 & BTCV \citeyearpar{landman2015miccai} [\href{https://www.synapse.org/#!Synapse:syn3193805/wiki/89480}{link}] & 47  & - & human \\
 & Pancreas-CT \citeyearpar{roth2015deeporgan} [\href{https://academictorrents.com/details/80ecfefcabede760cdbdf63e38986501f7becd49}{link}]            & 42  & - & human \\
 & CT-ORG \citeyearpar{rister2020ct} [\href{https://wiki.cancerimagingarchive.net/pages/viewpage.action?pageId=61080890#61080890cd4d3499fa294f489bf1ea261184fd24}{link}]& 140 & - & human \& AI \\
 & WORD \citeyearpar{luo2021word} [\href{https://github.com/HiLab-git/WORD}{link}] & 120 & - & human\\
 & LiTS \citeyearpar{bilic2019liver} [\href{https://competitions.codalab.org/competitions/17094}{link}]  & 130 & liver & human \\
 & AMOS22 \citeyearpar{ji2022amos} [\href{https://amos22.grand-challenge.org}{link}]  & 200 & - & human \& AI \\
 & KiTS \citeyearpar{heller2023kits21} [\href{https://kits-challenge.org/kits23/}{link}]  & 489 & kidney & human \\
 & AbdomenCT-1K \citeyearpar{ma2021abdomenct} [\href{https://github.com/JunMa11/AbdomenCT-1K}{link}]& 1,000 & - & human \& AI \\
 & MSD-CT \citeyearpar{antonelli2021medical} [\href{https://decathlon-10.grand-challenge.org/}{link}]   & 945 & liver, pancreas, colon & human \& AI \\
 & FLARE’23 \citeyearpar{ma2022fast} [\href{https://codalab.lisn.upsaclay.fr/competitions/12239}{link}]  & 4,100 & - & human \& AI \\
 & Abdominal Trauma Det \citeyearpar{rsna-2023-abdominal-trauma-detection} [\href{https://www.rsna.org/education/ai-resources-and-training/ai-image-challenge/abdominal-trauma-detection-ai-challenge}{link}]  & 4,711 & - & - \\
 
\midrule

\rowcolor{blue!10}\multicolumn{2}{l}{Private CT in \ourdataset} & 233  & liver, pancreas, kidney, colon, esophagus, uterus & human \& AI \\

\bottomrule
\end{tabular}

\caption{
\textbf{Dataset composition of \ourdataset.} Our \ourdataset\ comprises two components: CT scans from the public AbdomenAtlas 1.1 dataset and CT scans from a private source, totaling \numofscans\ tumor-annotated CT scans, with additional scans expected from various sources. Note that, for CT scans from AbdomenAtlas 1.1 dataset, we fully annotate six tumor types for each CT scan. 
}
\label{tab:public_dataset}
\end{table*}

\subsection{Comparative Models}
\label{sec:comparative_models_appendix}
The code for the Comparative Model is implemented in Python using MONAI and nnU-Net framework. 

\smallskip\noindent\textbf{\textit{nnU-Net Framework.}} nnU-Net serves as a framework for the automatic configuration of AI-driven semantic segmentation pipelines. When presented with a new segmentation dataset, it extracts pertinent metadata from the training cases to automatically determine its hyperparameters. It has withstood the test of time and continues to deliver state-of-the-art results. nnU-Net effectively illustrates that meticulously configuring and validating segmentation pipelines across a diverse range of segmentation tasks can yield a remarkably powerful algorithm. 

We implement UNETR, Swin UNETR, nnU-Net, ResEncM, and STU-Net using the nnU-Net framework. The orientation of CT scans is adjusted to specific axcodes. Isotropic spacing is employed to resample each scan, achieving a uniform voxel size of $1.5 \times 1.5 \times 1.5 mm^3$. Additionally, the intensity in each scan is truncated to the range [{-}175, 250] and then linearly normalized to [0, 1]. During training, we crop random fixed-sized $96 \times 96 \times 96$ regions, selecting centers from either a foreground or background voxel according to a pre-defined ratio. Furthermore, the data augmentation during training adheres to the default strategies outlined in the nnU-Net framework. All models are trained for 1000 epochs, with each epoch consisting of 250 iterations. Besides, we utilize the SGD optimizer with a base learning rate of 0.01, and the batch size is defined as 2. During inference, we utilize the test time augmentation by following the default implementations in nnU-Net framework. Besides, we use the sliding window strategy by setting the overlapping area ratio to 0.5.

\smallskip\noindent\textbf{\textit{MONAI Framework.}} MONAI (Medical Open Network for AI) is an open-source framework that supports AI in healthcare. Built on PyTorch, it offers a comprehensive set of tools for configuring, training, inferring, and deploying medical AI models. We implement SegResNet, Universal Model, and Suprem utilizing the MONAI framework. Since different methods have varying hyperparameter settings, we trained and tested the models exactly according to the original hyperparameters specified in the corresponding papers.

\clearpage

\clearpage
\section{Visual Real Examples in \ourdataset}
\label{sec:visual_real_examples_appendix}
\begin{figure}[h]
	\centering
	\includegraphics[width=\columnwidth]{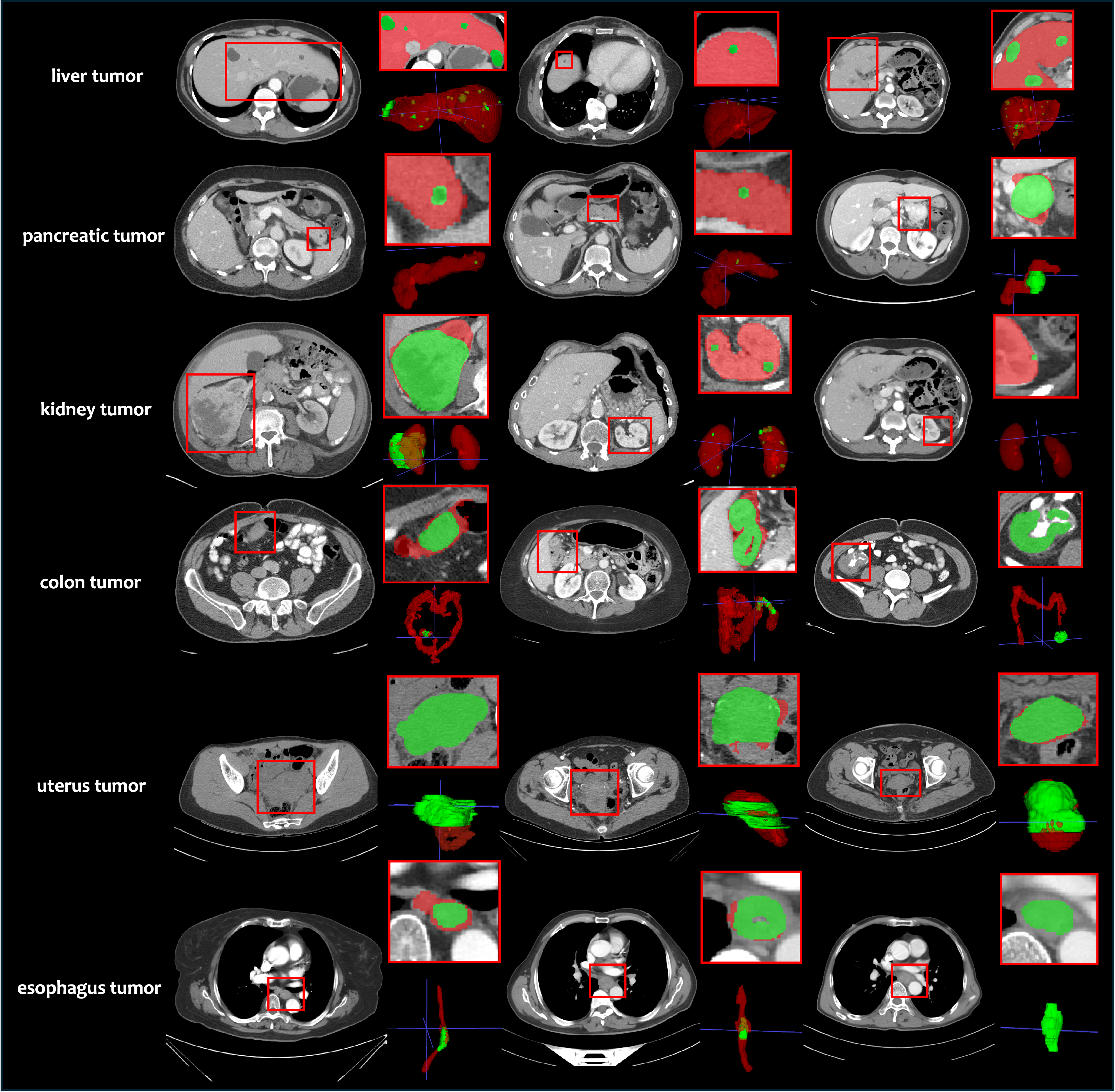}
    \caption{
    \textbf{Visual examples of six tumor types annotated in \ourdataset.} \ourdataset\ features a diverse distribution across various tumor stages and sizes. These comprehensive, high-quality tumors, accompanied by per-voxel annotations, significantly improve the performance of AI models, both on in-distribution and out-of- distribution data. (\figureautorefname~\ref{fig:scaling_analysis_with_real_data}).
    } 
	\label{fig:supp_ourdataset}
\end{figure}

\clearpage
\section{More Results: Best Lesson from Real Data}
\label{sec:more_results_best_lesson_from_real_appendix}
\begin{figure*}[h]
	\centering
\includegraphics[width=0.9\linewidth]{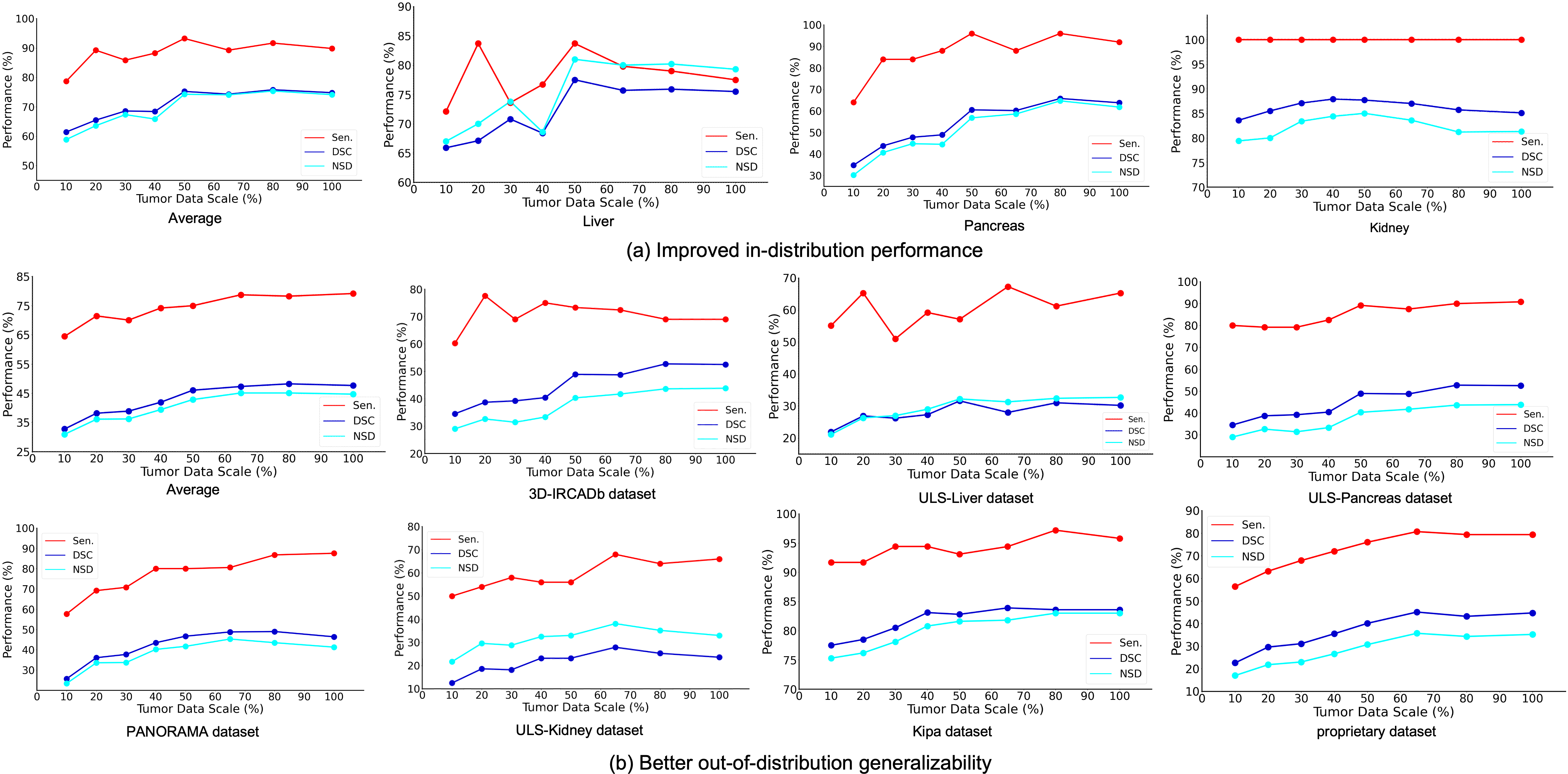}
	\caption{
    \textbf{Best Lesson from Real Data: Results on in-distribution and out-of-distribution data.} (a): while increasing data scale initially enhances in-distribution performance across various metrics (sensitivity, DSC, and NSD), it eventually plateaus. Notably, certain organ types, such as the Liver and Kidney, exhibit a decline in performance at the largest scales. (b): In contrast, the scaling trends observed in out-of-distribution datasets demonstrate consistent improvements in specific datasets (e.g., 3D-IRCADb, ULS-Pancreas) without reaching a plateau, indicating that larger data volumes may enhance generalizability. These results relate to the data-scaling lesson in \S\ref{sec:intro} (\upperbound\ if with real data only). Larger datasets are needed for effective out-of-distribution generalizability. 
    }
	\label{fig:scaling_analysis_with_real_data}
\end{figure*}

\clearpage
\section{More Results: Best Lesson for generalizability}
\label{sec:more_results_best_lesson_from_synt_appendix}

\begin{figure}[h]
	\centering
	\includegraphics[width=\columnwidth]{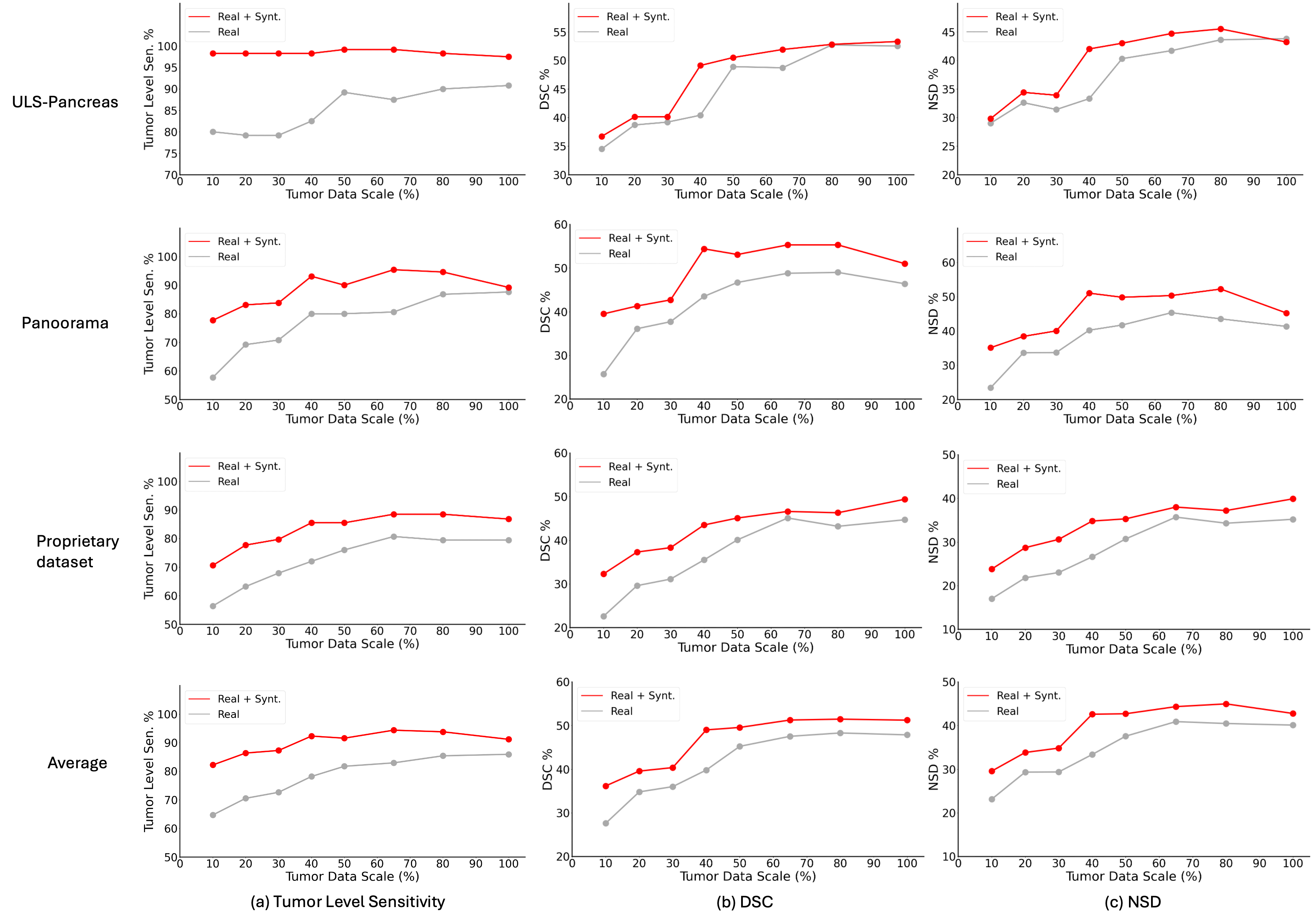}
    \caption{
    \textbf{Best lesson for pancreatic tumors.} Integrating real and synthetic data, compared to using real data alone, consistently improves generalizable performance in sensitivity, DSC, and NSD across various scenarios and data scales. These results underscore the benefits of this combination in enhancing the accuracy of pancreatic tumor analysis. 
    } 
	\label{fig:supp_best_lesson_pancreas_ood}
\end{figure}

\begin{figure}[h]
	\centering
	\includegraphics[width=\columnwidth]{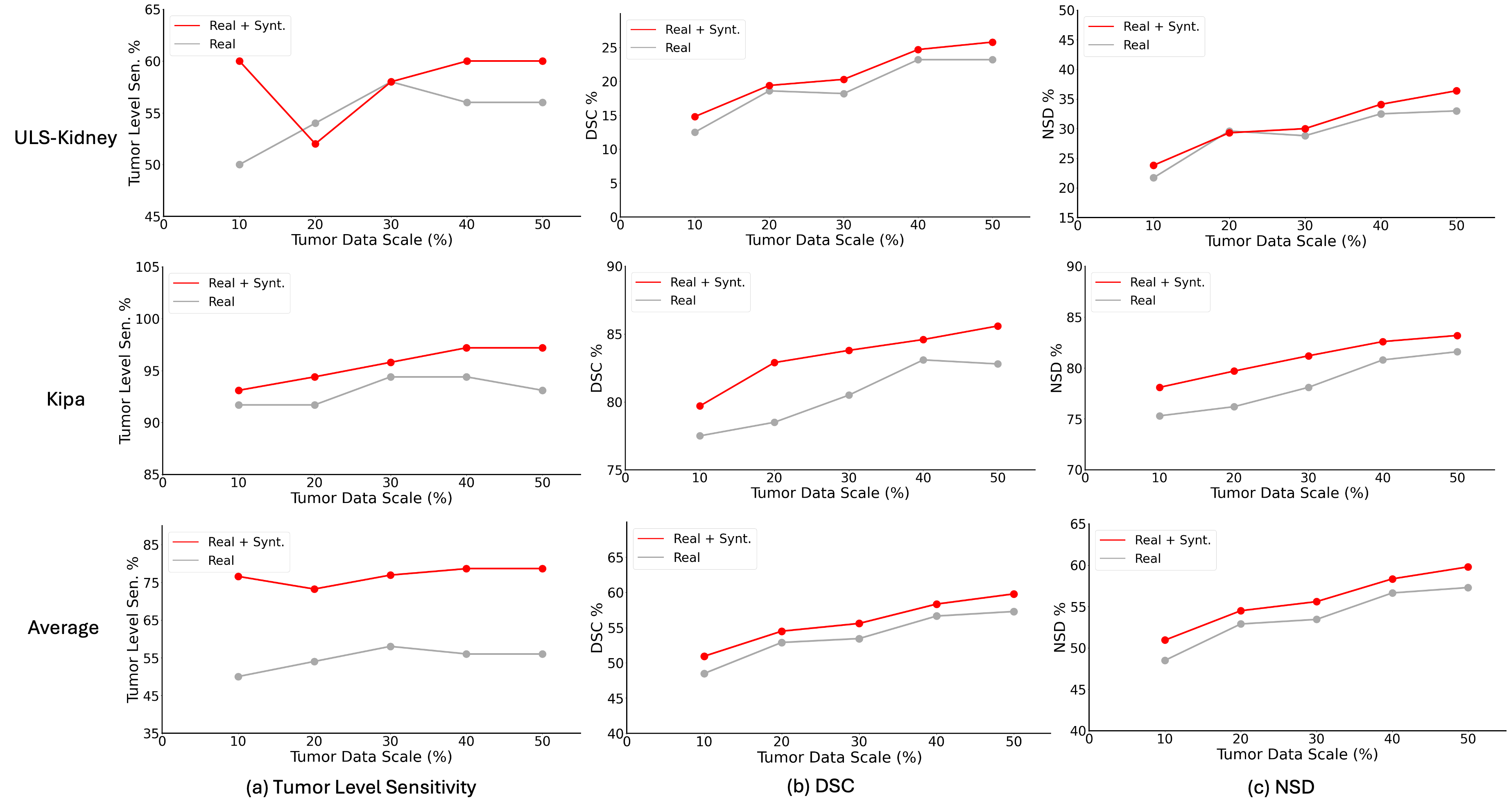}
    \caption{
    \textbf{Best lesson for kidney tumors.} Combining real and synthetic data consistently enhances generalizable performance in sensitivity, DSC, and NSD across various scenarios and data scales, highlighting its effectiveness in improving kidney tumor diagnosis accuracy.
    } 
	\label{fig:supp_best_lesson_kidney_ood}
\end{figure}
\clearpage

\end{document}